\documentclass[final]{cvpr}

\usepackage{times}
\usepackage{epsfig}
\usepackage{graphicx}
\usepackage{amsmath}
\usepackage{amssymb}

\usepackage{pifont}
\usepackage{booktabs}
\usepackage{multirow}
\usepackage{rotating}
\usepackage{subcaption}
\usepackage{algorithm}
\usepackage{algorithmic}

\graphicspath{{figures/}}

\newcommand{\Tref}[1]{Table~\ref{#1}}
\newcommand{\Eref}[1]{Eq.~(\ref{#1})}
\newcommand{\Fref}[1]{Fig.~\ref{#1}}

\newcommand{\Sref}[1]{Section~\ref{#1}}
\newcommand{\mymin}{\mathop{\rm min}\limits}

\newcommand{\cmark}{\ding{51}}%
\newcommand{\xmark}{\ding{55}}%

\newcommand*{\affmark}[1][*]{\textsuperscript{#1}}

\usepackage[pagebackref=true,breaklinks=true,colorlinks,bookmarks=false]{hyperref}

\def\cvprPaperID{4528} 
\def\confYear{CVPR 2021}

\begin{document}

\title{Divergence Optimization for Noisy Universal Domain Adaptation}

\author{Qing Yu\affmark[1,2]\ \ \ \ Atsushi Hashimoto\affmark[2]\ \ \ \ Yoshitaka Ushiku\affmark[2]\\
 \affmark[1]The University of Tokyo\ \ \ \ \affmark[2]OMRON SINIC X Corporation\\ 
{\tt\small yu@hal.t.u-tokyo.ac.jp}\ \ \ \ {\tt\small \{atsushi.hashimoto,yoshitaka.ushiku\}@sinicx.com}
}

\maketitle

\begin{abstract}
Universal domain adaptation (UniDA) has been proposed to transfer knowledge learned from a label-rich source domain to a label-scarce target domain without any constraints on the label sets. In practice, however, it is difficult to obtain a large amount of perfectly clean labeled data in a source domain with limited resources. Existing UniDA methods rely on source samples with correct annotations, which greatly limits their application in the real world. Hence, we consider a new realistic setting called Noisy UniDA, in which classifiers are trained with noisy labeled data from the source domain and unlabeled data with an unknown class distribution from the target domain. This paper introduces a two-head convolutional neural network framework to solve all problems simultaneously. Our network consists of one common feature generator and two classifiers with different decision boundaries. By optimizing the divergence between the two classifiers' outputs, we can detect noisy source samples, find ``unknown'' classes in the target domain, and align the distribution of the source and target domains. In an extensive evaluation of different domain adaptation settings, the proposed method outperformed existing methods by a large margin in most settings.
\end{abstract}

\section{Introduction}
Deep neural networks (DNNs) have achieved impressive results with large-scale annotated training samples, but the performance declines when the domain of the test data differs from the training data. To address this type of distribution shift between domains with no extra annotations, unsupervised domain adaptation (UDA) has been proposed to learn a discriminative classifier while there is a shift between training data in the source domain and test data in the target domain \cite{ben2010theory,  french2017self, ganin2015unsupervised, ghifary2016deep, saito2017asymmetric, saito2018maximum,  saito2018maximum, sener2016learning, taigman2016unsupervised, tzeng2017adversarial}.

Most existing domain adaptation methods assume that the source and target domains completely share the classes, but we do not know the class distribution of samples in the target domain in real-world UDA. Universal domain adaptation (UniDA) \cite{you2019universal} is proposed to remove the constraints on the label sets, where target samples may contain unknown samples belonging to classes that do not appear in the source domain and some source classes may not appear in the target samples. However, UniDA is still an ideal scenario, where existing UniDA methods require source samples with correct annotations to train the model. This requirement limits the application of existing UniDA methods in real domain adaptation problems, where clean and high-quality datasets are time consuming and expensive to collect. Data can more easily be collected from a crowd-sourcing platform or crawled from the Internet or social media, but such data are inevitably corrupted with noise (\textit{e}.\textit{g}. YFCC100M \cite{thomee2016yfcc100m}, Clothing1M \cite{xiao2015learning}, and ImageNet \cite{beyer2020we}). 

\begin{figure}[t]
    \centering
    \includegraphics[width=\linewidth]{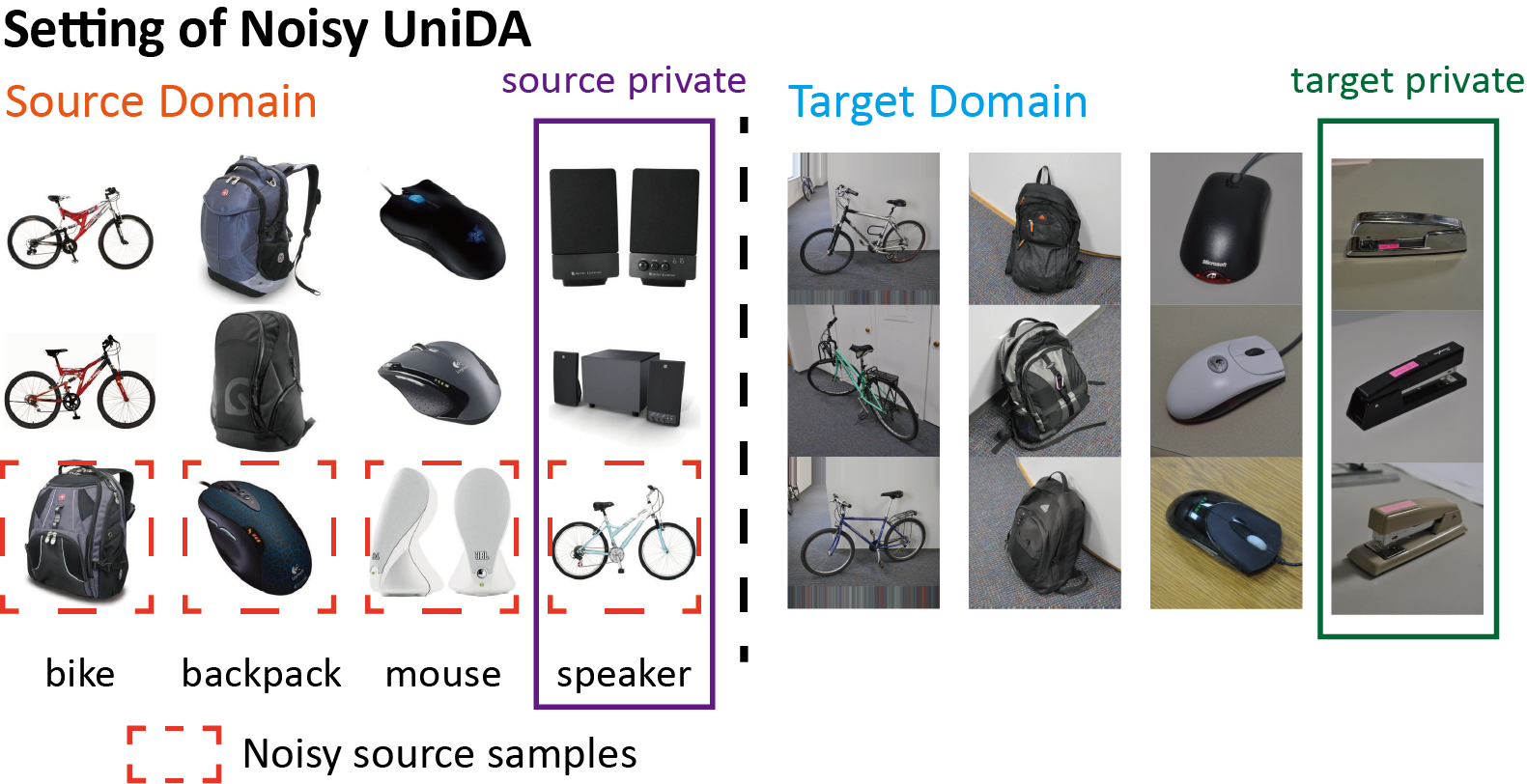}
    \caption{Problem setting of Noisy UniDA. Our proposed setting assumes that some source samples have corrupted labels, some classes of the source domain do not appear in the target domain, and the classes of some target samples are not shared by the source domain.}
    \label{fig:setting}
\end{figure}

Hence, we consider a new realistic setting called ``Noisy Universal Domain Adaptation'' (Noisy UniDA), as shown in \Fref{fig:setting}, which has the following properties:
\begin{itemize}
    \setlength{\parskip}{0mm}
	\setlength{\itemsep}{0mm}
    \item Labeled data of the source domain contains noisy labels. \footnote{The labels of target samples are not considered because they are not available in the setting of UDA.}
    \item Some classes of the source domain do not appear in the target domain, and these classes are named source private classes.
    \item Some classes of the target domain are not shared by the source domain, and these classes are named target private classes.
\end{itemize}

Some existing methods \cite{saito2018open,liu2019separate,shu2019transferable,cao2019learning, you2019universal} aim to solve certain parts of Noisy UniDA. For example, \cite{shu2019transferable} attempted to train domain-adaptive models on noisy source data, \cite{cao2019learning} worked on the partial problem that the source private classes are absent from the target domain, \cite{saito2018open} and \cite{liu2019separate} attempted to solve the open-set problem of target private classes, and \cite{you2019universal} addressed the settings with the partial problem and the open-set problem together. However, a method that can solve all these problems at the same time does not exist. 

Instead of solving each problem separately, we focus on the divergence of DNNs to address all the problems of Noisy UniDA. Inspired by Co-training for multi-view learning and semi-supervised learning \cite{blum1998combining, sindhwani2005co}, when different models having different parameters are trained on the same data, they learn distinct views of each sample because they have different abilities to learn. As a result, different models in each view would agree on the labels of most samples, and it is unlikely for compatible classifiers trained on independent views to agree on a wrong label. We find this property can be effective in Noisy UniDA, where the noisy source samples have wrong labels, and target private samples can also be considered to have incorrect labels because their true label is not contained in the label set. When these data are input to different networks, the networks are more likely to output different results because they have different parameters. Therefore, we utilize a two-head network architecture with two independent classifiers to detect all these unwanted samples simultaneously.

The proposed two-head network consists of one common feature generator and two separate label classifiers for classification. The two classifiers are updated by the same data at the mini-batch level, but they are initialized differently to obtain different classifiers. To detect noisy source samples in each mini-batch, we calculate the divergence between the two classifiers' outputs on the source data, and only source samples with small divergences are chosen to update the network by supervised loss. Using the same principle, target samples with larger divergence are more likely to be target private samples, and we further separate the divergence of the classifiers on common and target private samples to reject target private samples. Consequently, we align the distributions of the clean samples from the common classes shared by both domains, where the methods that align the entire distribution are influenced by incorrect source labels, the source private classes and target private classes.

We evaluated our method on a diverse set of domain adaptation settings. In many settings, our method outperforms existing methods by a large margin. We summarize the contributions of this paper as follows:
\begin{itemize}
    \setlength\itemsep{0pt}
    \item We propose a novel experimental setting and a novel training methodology for noisy universal domain adaptation (Noisy UniDA). 

    \item We propose a divergence optimization framework to detect noisy source samples, find target private samples, and align the distributions of the source and target domains according to the divergence of two label classifiers.

    \item We evaluate our method across several real-world domain adaptation tasks.
\end{itemize}

\section{Related Work}
Currently, there are several different approaches to UDA. \Tref{tbl:comp} summarizes the key methods.

One popular approach aims to match the distributions of the middle features in a convolutional neural network (CNN), and many such methods have been proposed \cite{bousmalis2016domain,ganin2016domain,purushotham2016variational,sun2016return,tzeng2014deep,saito2018maximum, long2018conditional}. A domain adversarial neural network (DANN) \cite{ganin2015unsupervised,ganin2016domain} and adversarial discriminative domain adaptation \cite{tzeng2017adversarial} introduced an adversarial training framework in which a domain discriminator is trained to distinguish two domains, while the feature extractor is trained to confuse the domain discriminator. Maximum classifier discrepancy (MCD) \cite{saito2018maximum} uses task-specific decision boundaries to align the source and target distributions.

\begin{table}[t]
\centering
\scalebox{1}{
\tabcolsep = 1.1mm
    \begin{tabular}{c|c|c|c}
        \toprule
         Method & Noisy labels & Partial DA & Open-set DA  \\                 \midrule
         DANN \cite{ganin2016domain} & \xmark   & \xmark & \xmark \\
         TCL \cite{shu2019transferable} & \cmark  & \xmark & \xmark  \\ 
         ETN \cite{cao2019learning} & \xmark    & \cmark & \xmark  \\ 
         STA \cite{liu2019separate} & \xmark & \xmark & \cmark  \\ 
         UAN \cite{you2019universal} & \xmark & \cmark & \cmark  \\ 
         DANCE \cite{saito2020dance} & \xmark & \cmark & \cmark  \\ 
         Proposed & \cmark   & \cmark & \cmark  \\
        \bottomrule
    \end{tabular}}
\caption{Summary of recent related methods. UniDA consists of Partial DA and Open-set DA. Our proposed method is the only method that covers all the settings.}
\label{tbl:comp}
\end{table}

Although these methods have achieved significant improvements, they all assume that the annotations of the source domain are clean, which is a limiting and expensive requirement in many real-world applications. Further, the source and target domains share the same classes, while the true class distribution of the target domain should be unknown. When these state-of-the-art domain adaptation methods are trained in a real-world setting, they may suffer from negative transfer owing to noisy source data and the class distribution of the target data, which degrades the generalization performance of the network.

The first problem of Noisy UniDA entails inaccurate annotations. There are studies on learning discriminative models from the datasets containing noisy labels \cite{reed2014training, zhang2016understanding, tanaka2018joint}. One strategy to reduce the effect of noise samples is updating the network only with samples having a small loss \cite{jiang2017mentornet,han2018co, yu2019does, wei2020combating}. Another approach uses robust loss functions \cite{patrini2017making}. For example, Zhang et al. \cite{zhang2018generalized} proposed the generalized cross-entropy loss, which is a generalization of the mean absolute error and the categorical cross-entropy. Other methods attempt to handle noisy source samples in a domain adaptation task. Transferable curriculum learning (TCL) \cite{shu2019transferable} uses the small-loss trick in DANN \cite{ganin2015unsupervised} to prevent the model from overfitting on noisy data.

The second problem of Noisy UniDA is when the classes of target samples are a subset of source classes, which is also referred to as partial domain adaptation. Studies in \cite{cao2018partial, zhang2018importance, cao2019learning} attempt to solve this task by finding the samples in the source domain that are similar to the target samples and place larger weights on these samples in the training process.

The last problem is the target private class of the target domain, that is, samples in the target domain that do not belong to a class in the source domain. To handle open-set recognition \cite{bendale2016towards} in a domain adaptation task, open-set domain adaptation by back-propagation \cite{saito2018open} trains a feature generator to lead the probability for the unknown class of a target sample to deviate from a predefined threshold. This approach trains the feature extractor and classifier in an adversarial training framework. A study of \cite{liu2019separate} used a coarse-to-fine separation pipeline to detect the unknown class and add one more class to the source classifier for the unknown class in an adversarial learning framework. Universal domain adaptation was proposed in \cite{you2019universal} and aims to handle partial domain adaptation and open-set domain adaptation at the sample time by importance weighting on both source samples and target samples. Domain adaptative neighborhood clustering via entropy optimization (DANCE) \cite{saito2020dance} also works on UniDA by using neighborhood clustering and entropy separation to achieve weak domain alignment.

In this study, we have developed a method to address all the mentioned problems of Noisy UniDA simultaneously. Specifically, the proposed method is robust against the noise levels of source sample annotations, the setting in which the target domain has a subset of source classes, and target private samples in the target domain.

\section{Method}
In this section, we present our proposed method for Noisy UniDA. First, we define the problem statement in \Sref{sec:problem}. Second, we illustrate the overall concept of the method in \Sref{sec:idea}. Then, our loss function is explained in \Sref{sec:loss}. Finally, we detail the actual training procedure in \Sref{sec:step}.

 \subsection{Problem Statement}
 \label{sec:problem}
 We assume that a source image-label pair $\{\boldsymbol{x_{s}},\boldsymbol{y_{s}}\}$ is drawn from a set of labeled source images, $\{X_{s}, Y_{s}\}$, while an unlabeled target image $\boldsymbol{x_{t}}$ is drawn from unlabeled images $X_{t}$. $\boldsymbol{y_{s}}$ is the one-hot vector of the class label $y_{s}$. We used $C_s$ and $C_t$ to denote the label sets of the source and target domains, respectively , and $C = C_s \cap C_t$ to represent the common label set shared by both domains. We also assume that the respective true labels for the source and target images are $Y^{GT}_{s}$ and $Y^{GT}_{t}$ ($y^{GT}_{s}$ and $y^{GT}_{t}$ for single source and target images, respectively), which implies that $y^{GT}_{s} \in C_s $ and $y^{GT}_{t} \in C_t$.
 For Noisy UniDA, we need to learn transferable features and train an accurate classifier across the source and target domains under the following conditions:
\begin{itemize}
    \setlength{\parskip}{0mm}
	\setlength{\itemsep}{0mm}
    \item The source image labels $Y_{s}$ are corrupted with noise, that is, $\exists \{\boldsymbol{x_{s}}, y_{s}\}$, $y_{s} \neq y^{GT}_{s}$.
    \item Some classes of the source domain do not appear in the target domain, that is, $C \subset C_s$. These source private classes are denoted by $\overline{C_s} = C_s \setminus C$.
    \item Target private samples exist in the target domain, that is, $C \subset C_t$. These target private classes are denoted by $\overline{C_t} = C_t \setminus C$.
\end{itemize}

Because the training process is performed at the mini-batch level, $D_s = \{(\boldsymbol{x_{s}^i}, \boldsymbol{y_{s}^i})\}^{N}_{i=1}$ is denoted as a mini-batch with size $N$ sampled from the source samples and $D_t = \{(\boldsymbol{x_{t}^i})\}^{N}_{i=1}$ is denoted as a mini-batch with size $N$ sampled from the target samples.

\subsection{Overall Concept}

\label{sec:idea}
To handle the problems of Noisy UniDA, we need to train the network to classify source samples correctly under the supervision of noisy labeled source samples and align the distribution of the source samples and the target samples, dealing with source private samples and target private samples simultaneously.

We focused on the divergence of DNNs, which is able to address all the problems of Noisy UniDA. Intuitively, since different classifiers can generate different decision boundaries and then have different abilities to learn, they learn distinct views of each sample, and the way that they are influenced by the noisy labels should also be different. As a result, different models are likely to agree on labels of most examples, and they are unlikely to agree on the incorrect labels of noisy training samples \cite{blum1998combining, sindhwani2005co}, which leads to a large divergence between the outputs of the networks. Therefore, it is not only possible to detect source samples that have wrong annotations, but it is also possible to identify target private samples, which can be considered to have incorrect annotations because their true label does not exist in the label set.

\begin{figure}[t]
    \centering
    \includegraphics[width=8cm]{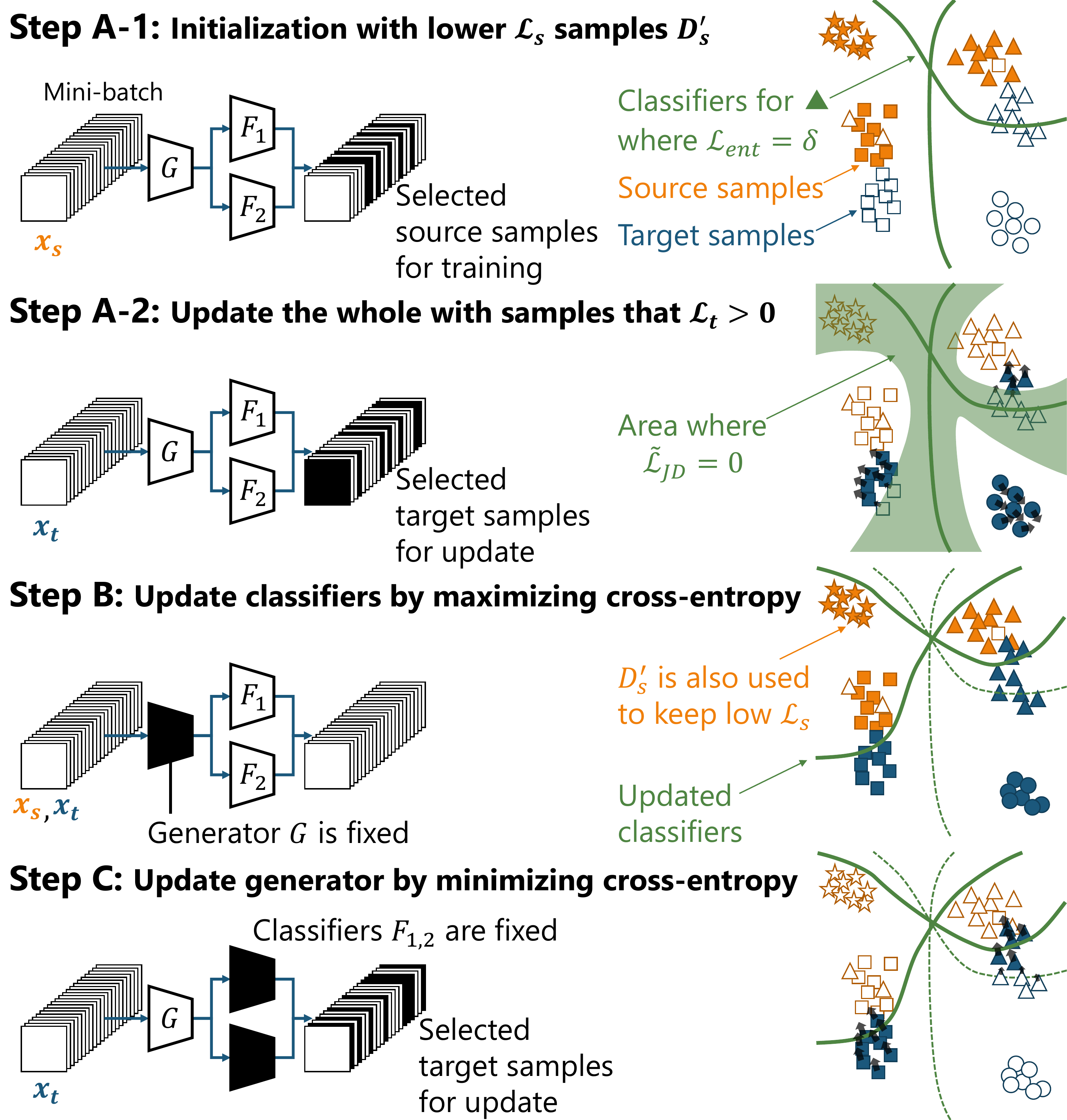}
    \caption{Training steps in the proposed method. There are four classes in the source domain and the target domain. Two of them are common to both domains. The white samples are not used for training, while the filled ones are used.}
    \label{fig:method_overview}
\end{figure}

To achieve that, we proposed a divergence optimization strategy utilizing a two-head CNN, as shown in \Fref{fig:method_overview}. The two-head CNN consists of a feature generator network $G$, which takes inputs $\boldsymbol{x_{s}}$ or $\boldsymbol{x_{t}}$, and two classifier networks, $F_1$ and $F_2$, which take features from $G$ and classify them into $|C_s|$ classes. The two classifiers are trained with the sames data at mini-batch level but they are initialized with random initial parameters. The classifier networks $F_1$ and $F_2$ output a $|C_s|$-dimensional vector of logits; then, the class probabilities can be calculated by applying the softmax function for the vector. The notations $\boldsymbol{p}_1(\boldsymbol{y}|\boldsymbol{x})$ and $\boldsymbol{p}_2(\boldsymbol{y}|\boldsymbol{x})$ denote the $|C_s|$-dimensional softmax class probabilities for input $\boldsymbol{x}$ obtained by $F_1$ and $F_2$, respectively. $p_1^k(\boldsymbol{y}|\boldsymbol{x^i})$ and $p_2^k(\boldsymbol{y}|\boldsymbol{x^i})$ represents the probability that samples $\boldsymbol{x^i}$ belong to the class $k$ predicted by each classifier.

To handle noisy labeled source samples, we calculated the divergence between the two classifiers' outputs for each source sample in the mini-batch. Because the two classifiers trained independently have different abilities to learn the noisy label, they tend to output similar predictions on clean samples and output different predictions on noisy samples. In addition to the popular small-loss technique to filter out noisy samples, we additionally selected samples with small divergences to update the network in each mini-batch. Similar to \cite{wei2020combating}, we further minimized the divergence of correct labeled source samples, which maximizes the agreement of the two classifiers, to achieve better results.

To address the problems of source private samples and target private samples, we propose a divergence separation loss for the target samples. Since target private samples can also be considered as noisy samples with incorrect labels, they will have larger divergences than target common samples. Therefore, by separating the divergences of the target samples, we can filter out some target private samples to achieve stable performance. Inspired by existing methods \cite{lee2019sliced, lee2019drop, saito2017adversarial, saito2018maximum} that utilize multiple classifiers with different parameters to achieve domain adaptation, we further use the two classifiers as a discriminator to detect target samples far from the support of the source domain. Then, we train the generator to minimize the divergence, thereby avoiding the generation of target features outside the support of the source.

However, in contrast to existing methods \cite{lee2019sliced, lee2019drop, saito2017adversarial, saito2018maximum} that align the entire distribution of the target domain with the distribution of the source domain, \textit{we select target samples having small divergences to update the feature generator to align the distribution partially}. By selecting samples with relatively small divergences to achieve partial alignment, we cannot only filter out target private classes to address the existence of the target private classes, but we can also focus on the samples exposed to the category boundaries to address the absence of the source private classes.

We also show the behavior of our method through a visualization of a toy problem in the supplementary.

\subsection{Symmetric Kullback-Leibler (KL) Divergence and Joint Divergence}
\label{sec:loss}
In \Sref{sec:idea}, we mentioned that the divergence between the two classifiers can be used to detect source samples with clean annotations and target samples in target private classes. The divergence we used is based on the symmetric KL divergence, which is defined by the following equation:
\begin{equation}
    \mathcal{L}_{SKLD}(D_s) = {\frac{1}{N}}\sum_{i=1}^{N}D_{\mathrm{KL}}(\boldsymbol{p}_1||\boldsymbol{p}_2)+{\frac{1}{N}}\sum_{i=1}^{N}D_{\mathrm{KL}}(\boldsymbol{p}_2||\boldsymbol{p}_1),
\end{equation}
where
\begin{equation}
    D_{\mathrm{KL}}(\boldsymbol{p}_1||\boldsymbol{p}_2)=\sum_{k=1}^{|C_s|}{p_1^k(\boldsymbol{y}|\boldsymbol{x^i_s}) \log {\frac{p_1^k(\boldsymbol{y}|\boldsymbol{x^i_s})}{p_2^k(\boldsymbol{y}|\boldsymbol{x^i_s})}}},
\end{equation}
\begin{equation}
    D_{\mathrm{KL}}(\boldsymbol{p}_2||\boldsymbol{p}_1)=\sum_{k=1}^{|C_s|}{p_2^k(\boldsymbol{y}|\boldsymbol{x^i_s}) \log {\frac{p_2^k(\boldsymbol{y}|\boldsymbol{x^i_s})}{p_1^k(\boldsymbol{y}|\boldsymbol{x^i_s})}}}.
\end{equation}
For source classes, we directly used $\mathcal{L}_{SKLD}$ to measure the agreement of the classifiers to detect the samples with clean annotations and minimized $\mathcal{L}_{SKLD}$ on these clean samples.

For target classes, when we attempted to use the divergence to detect the samples from the target private classes, we considered that $D_{\mathrm{KL}}(\boldsymbol{p}_1||\boldsymbol{p}_2)$ can be rewritten as follows:

\begin{eqnarray}
\begin{aligned}
    D_{\mathrm{KL}}(\boldsymbol{p}_1||\boldsymbol{p}_2)=&\sum_{k=1}^{|C_s|}{p_1^k(\boldsymbol{y}|\boldsymbol{x^i_t}) \log {p_1^k(\boldsymbol{y}|\boldsymbol{x^i_t})}}
    \\ &-
    \sum_{k=1}^{|C_s|}{p_1^k(\boldsymbol{y}|\boldsymbol{x^i_t}) \log {p_2^k(\boldsymbol{y}|\boldsymbol{x^i_t})}}
    \\ = -H(\boldsymbol{p}_1(\boldsymbol{y}|\boldsymbol{x_t}))&+H(\boldsymbol{p}_1(\boldsymbol{y}|\boldsymbol{x_t}),\boldsymbol{p}_2(\boldsymbol{y}|\boldsymbol{x_t})),
\end{aligned}
\end{eqnarray}
where $H(\boldsymbol{p}_1(\boldsymbol{y}|\boldsymbol{x_t}))$ is the entropy of $\boldsymbol{p}_1(\boldsymbol{y}|\boldsymbol{x_t})$ and $H(\boldsymbol{p}_1(\boldsymbol{y}|\boldsymbol{x_t}),\boldsymbol{p}_2(\boldsymbol{y}|\boldsymbol{x_t}))$ is the cross-entropy for $\boldsymbol{p}_1(\boldsymbol{y}|\boldsymbol{x_t})$ and $\boldsymbol{p}_2(\boldsymbol{y}|\boldsymbol{x_t})$.

Thus, $\mathcal{L}_{SKLD}$ can be rewritten as
\begin{eqnarray}
\begin{aligned}
    \label{eq:kl}
    \mathcal{L}_{SKLD}(D_t) &= {\frac{1}{N}}\sum_{i=1}^{N}\mathcal{L}_{crs}(D_t)-{\frac{1}{N}}\sum_{i=1}^{N}\mathcal{L}_{ent}(D_t)
    \\ \mathcal{L}_{crs}(D_t) &= H(\boldsymbol{p}_1(\boldsymbol{y}|\boldsymbol{x_t}),\boldsymbol{p}_2(\boldsymbol{y}|\boldsymbol{x_t}))\\&+H(\boldsymbol{p}_2(\boldsymbol{y}|\boldsymbol{x_t}),\boldsymbol{p}_1(\boldsymbol{y}|\boldsymbol{x_t}))
    \\\mathcal{L}_{ent}(D_t) &=H(\boldsymbol{p}_1(\boldsymbol{y}|\boldsymbol{x_t}))+H(\boldsymbol{p}_2(\boldsymbol{y}|\boldsymbol{x_t})),
\end{aligned}
\end{eqnarray}
where the first term shows the divergence of the two classifiers' outputs and the second term shows the entropy of each classifier output.

In \Sref{sec:idea}, we mentioned that target private samples are likely to have larger divergence than target common samples. If we directly use the symmetric KL divergence to measure the divergence, the class probabilities of target private samples should have ``small'' entropy owing to the second term in \Eref{eq:kl} having a minus symbol. However, because the target private samples do not belong to any source classes, the prediction confidence of these samples should be low, indicating that their class probabilities should have ``large'' entropy.

Thus, we modified the symmetric KL divergence to ``Joint Divergence'' as follows to detect target private samples:
\begin{equation}
\label{eq:kl_mod}
    \mathcal{L}_{JD}(D_t) = {\frac{1}{N}}\sum_{i=1}^{N}\mathcal{L}_{crs}(D_t) + {\frac{1}{N}}\sum_{i=1}^{N} \mathcal{L}_{ent}(D_t),
\end{equation}
where a larger divergence indicates a larger disagreement between the two classifiers and a lower confidence of each prediction. We further minimize \Eref{eq:kl_mod} for the detected target common samples and maximize it for the detected target private samples. The next section details the training procedure.

\subsection{Training Procedure}
\label{sec:step}
From previous discussions in \Sref{sec:idea} and \Sref{sec:loss}, we propose a training procedure consisting of the following three steps, as shown in \Fref{fig:method_overview}. \textit{The three steps are repeated at the mini-batch level in our method.}

\textbf{Step A-1}
First, we trained the entire network containing both classifiers and generator to learn discriminative features and classify the source samples correctly under the supervision of labeled source samples. Because small loss samples are likely to have correct labels \cite{han2018co, wei2020combating}, we trained our classifier using only small-loss instances in each mini-batch data to make the network resistant to noisy labels. For the loss function, cross-entropy loss is commonly used, which is denoted as follows:
\begin{equation}
\begin{aligned}
   \mathcal{L}_{sup}(D_s) = &-{\frac{1}{N}}\sum_{i=1}^{N}\sum_{k=1}^{|C_s|}y_s^i\log p_1^k({\boldsymbol y}|{\boldsymbol {x^i_s}})
   \\&-{\frac{1}{N}}\sum_{i=1}^{N}\sum_{k=1}^{|C_s|}y_s^i\log p_2^k({\boldsymbol y}|{\boldsymbol {x^i_s}}).
\end{aligned}
\end{equation}

As mentioned in \Sref{sec:loss}, we also added the agreement of the two classifiers to the loss for selecting clean samples. Therefore, the loss on source samples is expressed as follows:
\begin{equation}
\label{eq:Ls}
\begin{aligned}
   \mathcal{L}_{s}(D_s) = \mathcal{L}_{sup}(D_s) +\lambda \mathcal{L}_{SKLD}(D_s),
\end{aligned}
\end{equation}
where $\lambda$ is a hyperparameter and $\lambda$ is set to 0.1 in all the experiments.
To filter out noisy samples, we used the joint loss \Eref{eq:Ls} to select small loss samples because a noisy sample is more likely to have larger cross-entropy loss and larger divergence. Specifically, we conducted small-loss selection as follows:
\begin{equation}
\label{eq:select}
D_s'={\arg\min}_{D':|D'|\ge \alpha|D_s|}\mathcal{L}_{s}(D_s).
\end{equation}
This equation indicates that we only use $\alpha\%$ samples in a mini-batch to the network. The objectives are as follows:

\begin{equation}
\label{eq:min_ls}
  \mymin_{G,F_1,F_2} \mathcal{L}_s(D_s').
\end{equation}

\textbf{Step A-2}
In addition to the supervised training on source samples, we attempt to detect target private samples using $\mathcal{L}_{JD}$ from \Eref{eq:kl_mod}. To increase $\mathcal{L}_{JD}$ for target private samples and decrease it for common samples, we introduce a threshold $\delta$ and a margin $m$ to separate the $\mathcal{L}_{JD}$ on target samples, using the following equations:

\begin{equation}
\label{eq:Lt}
\begin{aligned}
   &\mathcal{L}_{t}(D_t) = \mathcal{\tilde{L}}_{JD}(D_t)={\frac{1}{N}}\sum_{i=1}^{N}\mathcal{\tilde{L}}_{crs}(D_t)+{\frac{1}{N}}\sum_{i=1}^{N}\mathcal{\tilde{L}}_{ent}(D_t)
   \\&\mathcal{\tilde{L}}_{crs}(D_t)=
   \begin{cases}
   -|\mathcal{L}_{crs}(\boldsymbol {x_t})-\delta| & \text{if } |\mathcal{L}_{crs}(\boldsymbol {x_t})-\delta| > m\\
   0 & \text{otherwise}
   \end{cases}
   \\&\mathcal{\tilde{L}}_{ent}(D_t)=
   \begin{cases}
   -|\mathcal{L}_{ent}(\boldsymbol {x_t})-\delta| & \text{if } |\mathcal{L}_{ent}(\boldsymbol {x_t})-\delta| > m\\
   0 & \text{otherwise}
   \end{cases}.
\end{aligned}
\end{equation}
Only when the divergence or the entropy of the two classifiers is larger or smaller enough, we further increased or decreased them to separate private or common target samples. The objectives are as follows:

\begin{equation}
  \mymin_{G,F_1,F_2} \mathcal{L}_{t}(D_t).
\end{equation}

Moreover, $D_t'$ is denoted as the detected target common samples with small divergences, which is $\{\boldsymbol{x_{t}^i}: \boldsymbol{x_{t}^i} \in D_t, \mathcal{L}_{crs}(\boldsymbol{x_{t}^i}) < \delta - m\} $.

\textbf{Step B}
Then, to align the distribution of the source and target domains, we trained the classifiers as a discriminator for a fixed generator to increase the divergence to make the network detect target samples that do not have the support of source samples (\textbf{Step B} in \Fref{fig:method_overview}). In this step, we also use source samples to reshape the support. The objective is as follows:

\begin{equation}
\label{eq:step_b}
  \mymin_{F_1,F_2} \mathcal{L}_s(D_s') - {\frac{1}{N}}\sum_{i=1}^{N}\mathcal{L}_{crs}(D_t). \\
\end{equation}
  
\textbf{Step C}
Finally, we trained the generator to minimize the divergence for fixed classifiers (\textbf{Step C} in \Fref{fig:method_overview}) to partially align the distributions of the target and source domains using the detected target common samples $D_t'$. The final objective is as follows:
\begin{equation}
\label{eq:step_c}
 \mymin_{G} {\frac{1}{N}}\sum_{i=1}^{N}\mathcal{L}_{crs}(D_t'). \\
 \end{equation}
This step is repeated $n$ times to achieve better alignment, and we set $n=4$ in all the experiments.

\subsection{Inference}
\label{sec:test}
    At inference time, to distinguish between common samples and target private samples, we considered the cross entropy $\mathcal{L}_{crs}$ between the two classifiers' outputs. When the divergence is above a detection threshold $\delta$, we assigned the sample as a target private sample, denoted by
    \begin{equation}
        \mathcal{L}_{crs}(\boldsymbol{x}) > \delta.
\end{equation}

\section{Experiment}
\label{sec:exp}
\subsection{Experimental Setup}
\textbf{Datasets.} Following previous studies \cite{you2019universal}, we used three datasets in the experiments. Office \cite{saenko2010adapting}, which has 3 domains
(Amazon, DSLR, Webcam) and 31 classes, was used as the first dataset. The second dataset is OfficeHome \cite{venkateswara2017deep} containing 4 domains (i.e., Art, Clipart, Product, and Real) and 65 classes. The last dataset is VisDA \cite{peng2017visda} containing two domains (i.e., synthetic and real) and 12 classes.
To construct the setting of Noisy UniDA, we split the class of each dataset as in \cite{you2019universal}. $|C|/|\overline{C_s}|/|\overline{C_t}|=10/10/11$ for Office, $10/5/50$ for OfficeHome, and $6/3/3$ for VisDA. We also used a noise transition matrix $Q$ to corrupt the source datasets manually \cite{han2018co,jiang2017mentornet} to simulate noisy source samples. We implemented two variations of $Q$ for (1) pair flipping and (2) symmetry flipping \cite{han2018co}. The noise rate $\rho$ was chosen from $\{0.2,0.45\}$. Intuitively, if $\rho=0.45$, almost half of the noisy source data was annotated with incorrect labels that cannot be learned without additional assumptions. In contrast, $\rho=0.2$ implied that only $20\%$ of labels were corrupted, which is a low-level noise situation. Note that pair flipping is much harder than symmetry flipping \cite{han2018co}. For each adaptation task, there were four types of noisy source data: \emph{Pair-}$20\%$ (P$20$), \emph{Pair-}$45\%$ (P$45$), \emph{Symmetry-}$20\%$ (S$20$),  and \emph{Symmetry-}$45\%$ (S$45$).

\begin{table*}
\begin{minipage}{0.58\linewidth}
\centering
\scalebox{0.8}{
\tabcolsep = 0.7mm
\begin{tabular}{c|cccc|cccc|cccc}
\toprule
\multirow{2}{*}{Method} & \multicolumn{4}{c|}{Office} & \multicolumn{4}{c|}{OfficeHome} & \multicolumn{4}{c}{VisDA} \\ \cline{2-13}
                        & P20   & P45  & S20  & S45  & P20    & P45   & S20   & S45   & P20  & P45  & S20  & S45  \\ \hline\hline
SO                      &   77.23    &  50.88    &  78.09    &  53.15    &    63.33    &   39.46    &  64.09     &   44.99    &   44.57   &   38.41   &  26.51    &   15.39   \\
TCL                     &   80.82    &  50.48    & 82.97     &  74.86    &   62.31     &  40.24     & 63.41      &   49.69    &  62.96    &   43.31   &  61.26    &   52.96   \\
ETN                     &   84.46    &  53.23    &  85.40    &  83.53    &    67.93    &  44.55     &  68.99     &   56.05    &   58.99   &   44.36   &  62.17    &   55.83   \\
STA                     &   83.12    &   54.74   &  83.19    & 68.27     &    64.31    &  44.22     &  65.53     &    49.04   &    41.62  &  41.50    &  52.17    &  42.32    \\
UAN                     &   72.39    &  45.64    &   77.59   &  64.85    &    70.90    &  41.31     &  73.79     &    59.67   &   53.93   &  42.60    &  53.25    &   47.70   \\
DANCE                   &   84.88    &  55.40     &  83.00    &  56.02    &    \textbf{77.32}    &  47.51     &   77.54    &   66.96    &   57.38   &  41.45    &  24.30    &   14.94   \\
DANCE$_{\text{sel}}$   &   86.07    &  56.32     &  91.24    &  79.82  &  76.49    &  48.45     &   \textbf{78.71}    &  64.17 &  63.93   &  43.91    &  62.97    &   52.32   \\ \hline
Ours                    &   \textbf{91.22}    &  \textbf{62.49}    &  \textbf{91.40}    &  \textbf{87.92}    &   76.10     &  \textbf{51.93}     &   77.46    &  \textbf{71.97}     &   \textbf{67.27}   &  \textbf{48.25}    &   \textbf{70.53}   & \textbf{57.82}  \\ \bottomrule
\end{tabular}
}
\caption{Average target-domain accuracy (\%) of each dataset under different noise types. We report the average accuracy over all tasks for each dataset. Bold values represent the highest accuracy in each row.}
\label{tbl:uni_results}
\end{minipage}
\hfill
\begin{minipage}{0.4\linewidth}
		\centering
		\includegraphics[width=0.87\textwidth]{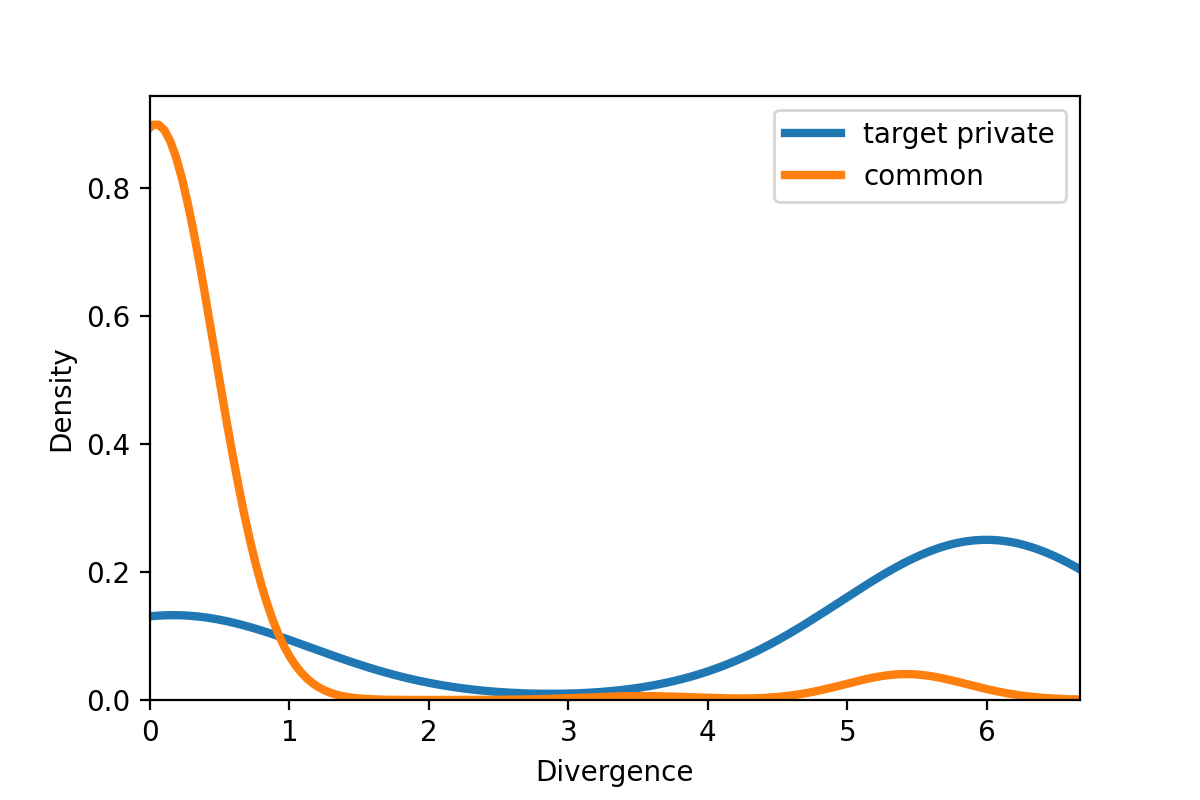}
	    \vspace{-10pt}
		\captionof{figure}{Probability density function of the divergence of common and target private samples (estimated by Gaussian kernel with Scott's rule ).}
		\label{fig:dis}
	\end{minipage}
\vspace{-10pt}
\end{table*}

\begin{table*}[t]
\centering
\scalebox{0.86}{
\tabcolsep = 0.7mm
\begin{tabular}{c|cccccc|c|cccccccccccc|c}
\multicolumn{21}{c}{Noise Type: P45}  \\
\toprule
\multirow{2}{*}{Method} & \multicolumn{6}{c|}{Office}        &     & \multicolumn{12}{c|}{OfficeHome}                                       &     \\ \cline{2-21}
                        & A2W & D2W & W2D & A2D & D2A & W2A & Avg & A2C & A2P & A2R & C2A & C2P & C2R & P2A & P2C & P2R & R2A & R2C & R2P & Avg \\ \hline\hline
SO &47.12&56.50&59.13&50.67&56.00&35.89&50.88&33.11&41.16&57.84&30.01&37.50&41.86&39.30&26.03&50.57&40.81&28.51&46.80&39.46     \\
TCL  &50.29&54.97&52.85&49.58&50.77&44.41&50.48&29.40&42.62&51.52&31.73&39.60&42.79&50.39&28.20&53.34&34.65&30.27&48.34&40.24    \\
ETN    &52.01&52.94&49.47&52.79&60.33&51.82&53.23&36.36&54.08&\textbf{65.48}&35.36&38.74&49.39&47.04&27.04&57.09&41.50&33.35&49.15&44.55    \\
STA  &\textbf{61.18}&53.42&54.19&55.93&63.05&40.69&54.74&32.21&42.02&60.62&38.38&42.37&53.13&50.29&30.48&58.99&42.47&28.60&51.03&44.22   \\
UAN &44.93&57.68&44.00&41.09&40.50&45.64&45.64&31.53&44.44&46.64&40.16&44.04&47.48&41.42&34.38&54.81&38.61&30.06&42.15&41.31  \\
DANCE  &46.80&56.82&53.85&56.17&\textbf{69.98}&48.80&55.40&36.10&39.69&63.12&39.62&41.60&46.84&57.04&32.28&\textbf{68.55}&50.26&39.82&55.20&47.51    \\
DANCE$_{\text{sel}}$  &60.35&60.32&\textbf{63.55}&48.79&54.42&50.48&56.32&34.32&41.95&64.19&47.25&42.24&\textbf{61.73}&46.32&40.92&60.40&\textbf{53.30}&33.86&54.87&48.45  \\ \hline
Ours  &58.93&\textbf{72.49}&56.06&\textbf{58.71}&65.86&\textbf{62.90}&\textbf{62.49}&\textbf{37.12}&\textbf{57.73}&54.17&\textbf{52.39}&\textbf{47.26}&55.22&\textbf{57.93}&\textbf{43.97}&64.17&50.36&\textbf{41.29}&\textbf{61.61}&\textbf{51.93} \\   \bottomrule
\multicolumn{21}{c}{Noise Type: S45}  \\
\toprule
\multirow{2}{*}{Method} & \multicolumn{6}{c|}{Office}        &     & \multicolumn{12}{c|}{OfficeHome}                                       &     \\ \cline{2-21}
                        & A2W & D2W & W2D & A2D & D2A & W2A & Avg & A2C & A2P & A2R & C2A & C2P & C2R & P2A & P2C & P2R & R2A & R2C & R2P & Avg \\ \hline\hline
SO &37.16&53.17&76.00&50.49&41.97&60.14&53.15&30.36&46.40&59.20&48.84&46.50&57.94&41.20&29.09&56.49&40.56&30.39&52.90&44.99    \\
TCL  &77.62&64.50&82.96&81.86&66.35&75.84&74.86&30.08&44.22&46.66&49.31&54.73&57.03&52.26&40.72&69.14&48.61&40.41&63.16&49.69    \\
ETN   &83.11&\textbf{80.64}&88.64&87.35&77.18&84.27&83.53&39.96&62.37&77.07&61.08&53.51&70.17&47.70&39.42&68.93&49.77&37.32&65.33&56.05    \\
STA  &60.94&70.22&88.80&71.6&44.26&73.79&68.27&37.65&50.75&57.14&51.35&53.64&67.39&39.92&32.09&64.29&47.19&32.63&54.49&49.04  \\
UAN &62.59&72.88&75.82&56.97&61.94&58.90&64.85&42.67&59.04&57.20&57.29&64.93&70.56&64.45&49.09&72.53&55.98&47.26&75.08&59.67  \\
DANCE  &21.72&69.45&85.76&27.68&51.04&80.47&56.02&38.29&\textbf{74.72}&87.24&\textbf{75.01}&\textbf{81.27}&80.20&55.57&44.05&78.08&61.59&47.70&79.83&66.96    \\
DANCE$_{\text{sel}}$ &82.45&69.16&92.08&86.53&75.51&73.22&79.82&40.33&58.27&73.80&61.15&67.76&78.36&70.63&\textbf{54.23}&78.53&63.22&48.21&75.56&64.17 \\\hline
Ours &\textbf{87.63}&76.87&\textbf{98.32}&\textbf{89.43}&\textbf{84.49}&\textbf{90.78}&\textbf{87.92}&\textbf{46.30}&69.52&\textbf{87.79}&70.34&70.55&\textbf{81.77}&\textbf{74.72}&54.15&\textbf{88.23}&\textbf{78.04}&\textbf{61.22}&\textbf{80.98}&\textbf{71.97} \\   \bottomrule
\end{tabular}
}
\vspace{-5pt}
\caption{Results on noisy universal domain adaptation of each task for Office and OfficeHome. Bold values represent the highest accuracy in each row.}
\vspace{-5pt}
\label{tbl:detail_results}
\end{table*}

\textbf{Compared Methods.} We compared the proposed method with (1) CNN: source only ResNet-50 (SO) \cite{he2016deep}, (2) label noise-tolerant domain adaptation method: TCL \cite{shu2019transferable}, (3) partial domain adaptation method: example transfer network (ETN) \cite{zhang2018importance}, (4) open-set domain adaptation method: separate to adapt (STA) \cite{liu2019separate}, and (5) universal domain adaptation methods: universal adaptation network (UAN) \cite{you2019universal}, DANCE \cite{saito2020dance}. Because these methods achieved state-of-the-art performance in their respective settings, it is valuable to show their performance in the Noisy UniDA setting. We also tried to incorporate the ``select'' operation in \Eref{eq:select} and \Eref{eq:min_ls} into the supervised source loss of DANCE to create a stronger baseline as DANCE$_{\text{sel}}$.

\textbf{Evaluation Protocols.} The same evaluation metrics as previous works are used, where the accuracy is averaged over $|C|+1$ classes and all the samples belonging to the target private classes are regarded as one unified unknown class. For example, when Office is used as the dataset, an average of 11 classes is reported. For methods that do not detect target private samples originally, we used confidence thresholding to reject target private samples.

\textbf{Implementation Details.} In this experiment, we used the same CNN architecture and hyperparameters as in \cite{saito2020dance}. We implemented our network based on ResNet-50 \cite{he2016deep}. We used the modules of ResNet until the {\it average-pooling} layer just before the last {\it fully-connected} layer as the generator and one {\it full-connected} layer as the classifier. We defined the threshold $\delta = \log |C_s|$ because $2 \times \log |C_s|$ is the maximum value of $H(\boldsymbol{p_1})+H(\boldsymbol{p_2})$, and the margin $m=1$ in all the experiments. The detailed analysis of sensitivity to hyper-parameters is discussed in the supplementary.
    
\subsection{Experimental Results}
\label{sec:result}
\Tref{tbl:uni_results} summarizes the results, which compare the proposed method with other state-of-the-art methods. Because the pair flipping noise is more difficult than the symmetry flipping noise, the accuracy under the setting of pair flipping noise is lower. However, \Tref{tbl:uni_results} clearly shows that our approach outperformed the existing methods in all the noise settings, as they could not handle all the difficulties of Noisy UniDA. ETN shows satisfactory results in the Office dataset, and TCL achieves high performance in the VisDA dataset, but our proposed method outperformed them with a considerable margin. In the difficult OfficeHome dataset, DANCE achieves high performance when the noise rate is low (e.g., P20 and S20). However, our method's performance is competitive with DANCE. When the noise rate increases (e.g., P45 and S45), the proposed method performs better. Adding the ``select'' operation to DANCE achieves some improvements in many settings comparing with the original DANCE, but our method still achieves better performance than DANCE$_{\text{sel}}$ in most settings.

\Tref{tbl:detail_results} shows the results of each task in each dataset when the noise type is P45 and S45. In most tasks, our method achieves the highest performance, showing its power to detect clean source samples, align the distribution partially, and discriminate the target private classes.

We also plot the probability density function (calculated by kernel density estimation) of the divergence $\mathcal{L}_{crs}$ of common and target private classes in the target domain in \Fref{fig:dis}, proving that the divergence between the two classifiers separates the common classes $C$ and the target private classes $\overline{C_t}$ well.

\begin{figure*}[t]
     \centering
     \begin{subfigure}[b]{0.33\textwidth}
         \centering
         \includegraphics[width=1.1\textwidth]{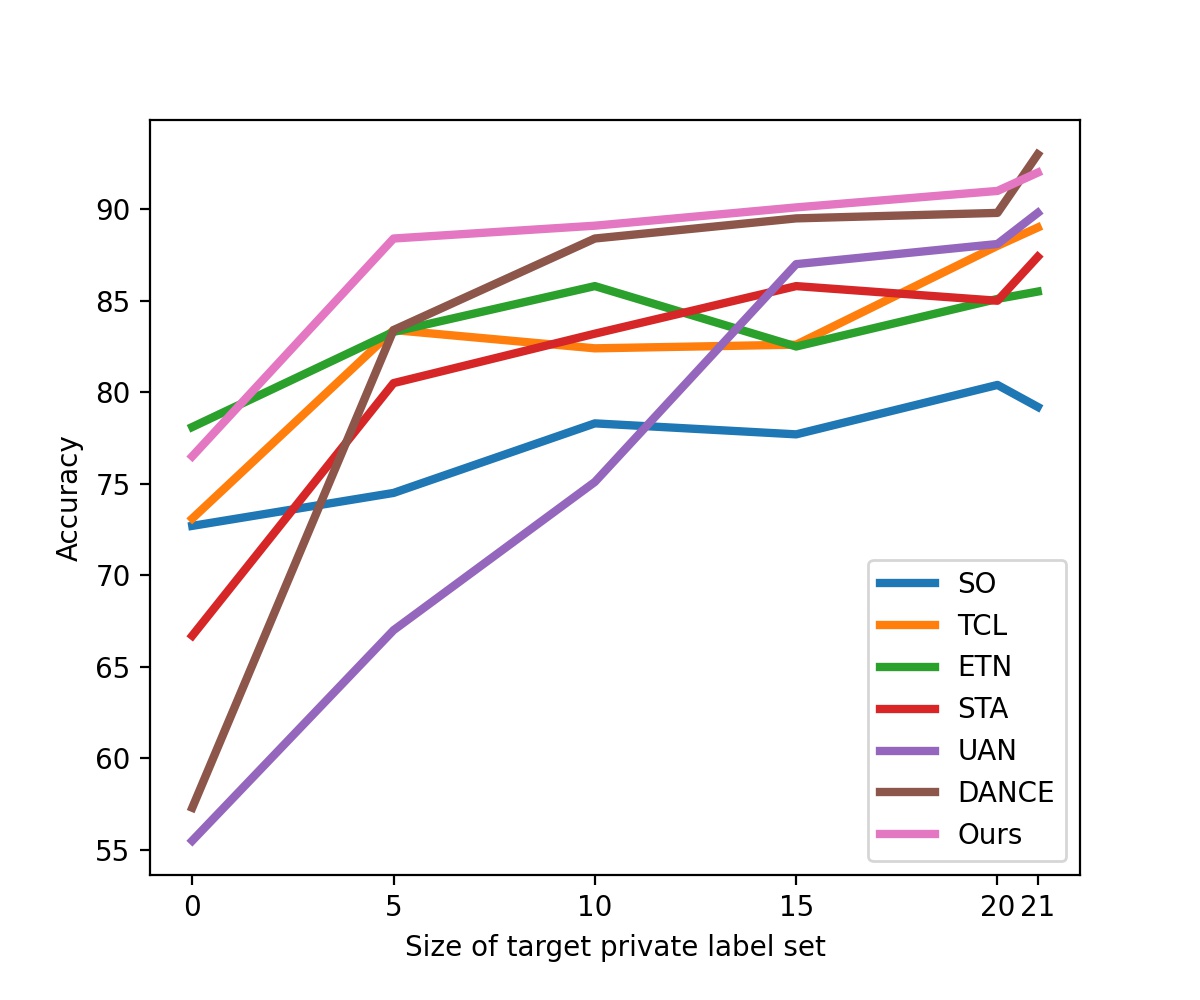}
         \caption{Accuracy w.r.t. $\overline{C_t}$.}
         \label{fig:ab_ct}
     \end{subfigure}
     \hfill
     \begin{subfigure}[b]{0.33\textwidth}
         \centering
         \includegraphics[width=1.1\textwidth]{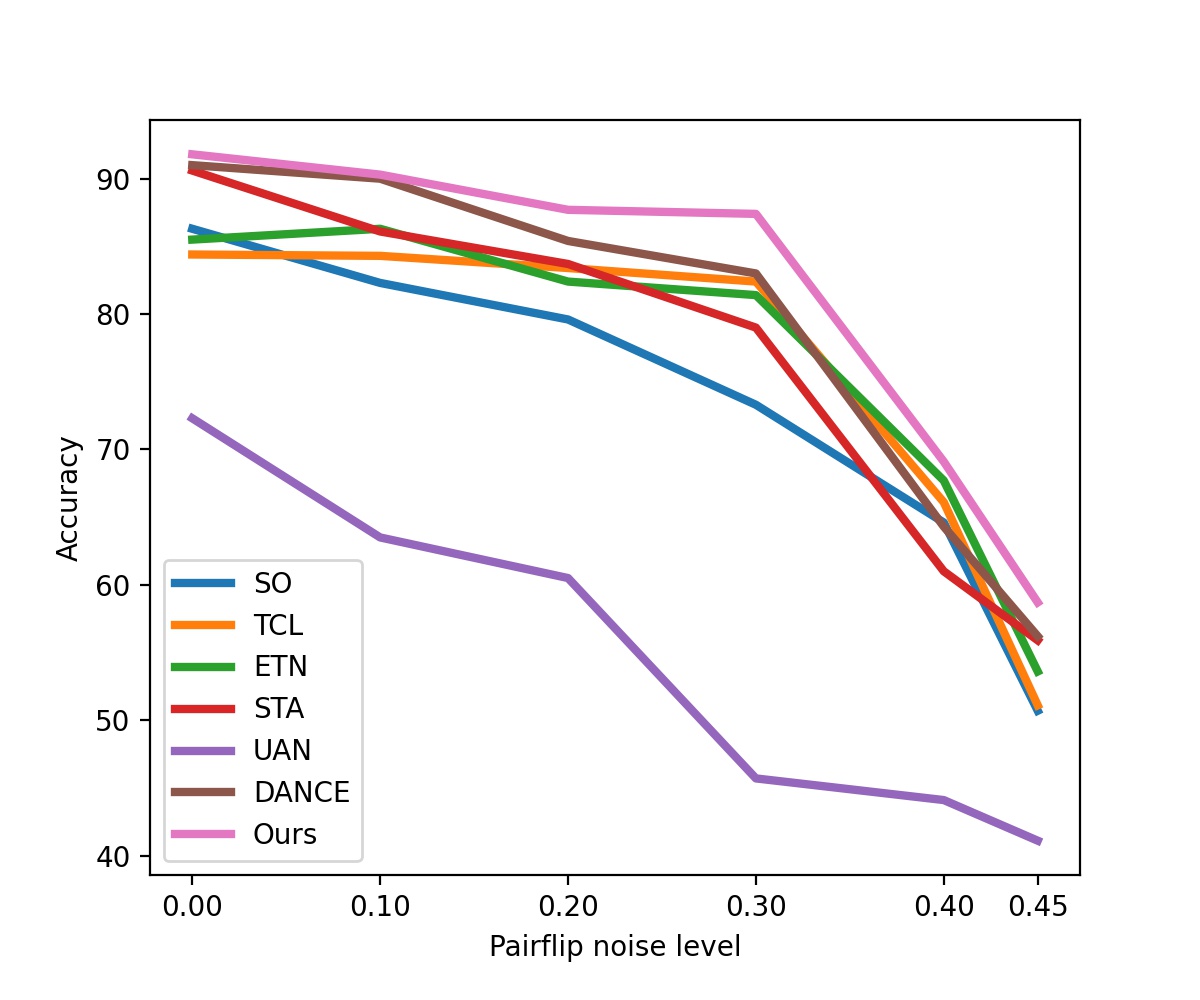}
         \caption{Accuracy w.r.t. pairflip noise level.}
         \label{fig:ab_p}
     \end{subfigure}
     \hfill
     \begin{subfigure}[b]{0.33\textwidth}
         \centering
         \includegraphics[width=1.1\textwidth]{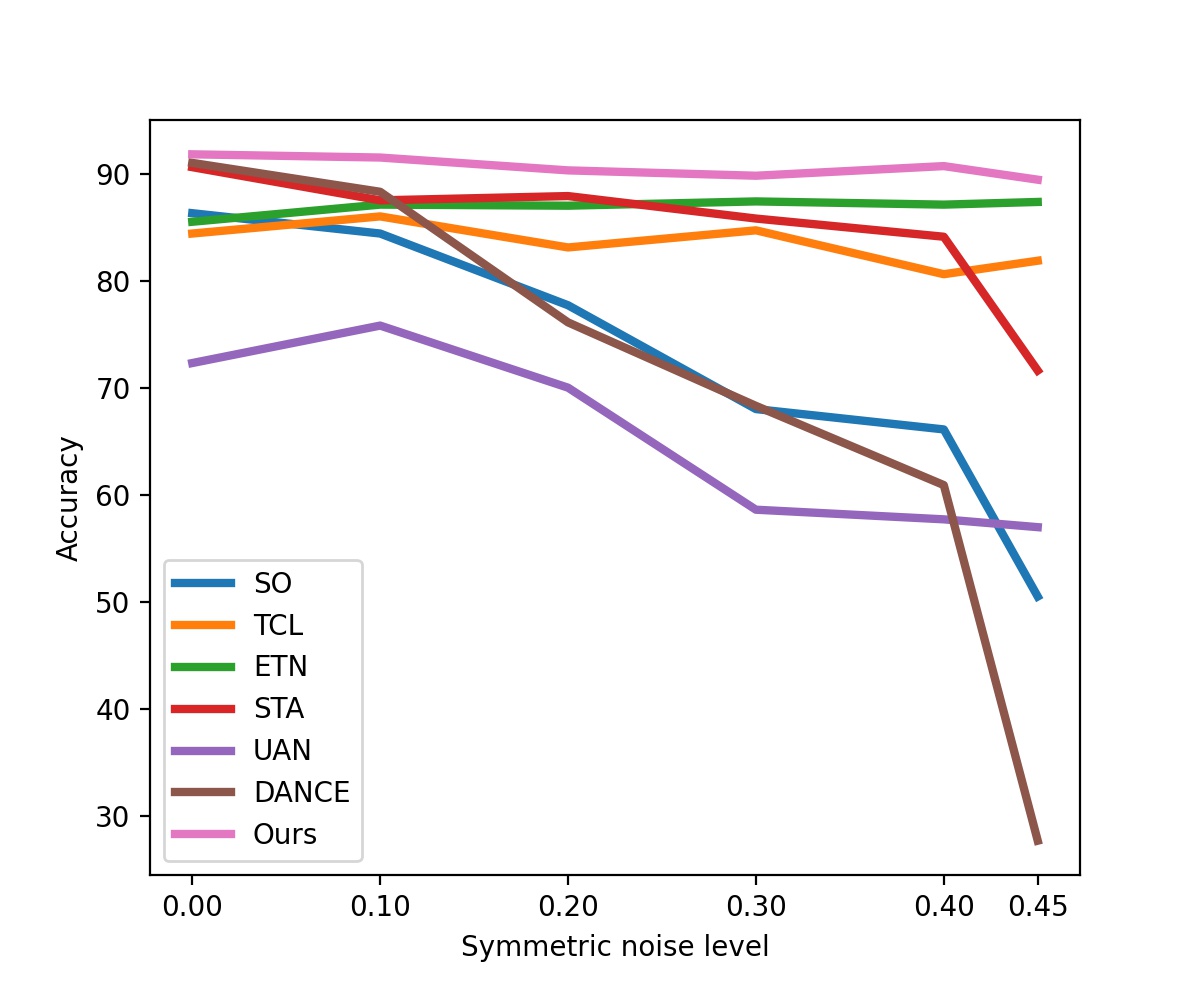}
         \caption{Accuracy w.r.t. symmetric noise level.}
        \label{fig:ab_s}
     \end{subfigure}
     \vspace{-15pt}
    \caption{Analysis of the performance in task A$\rightarrow$D under different Noisy UniDA settings.}
    \vspace{-10pt}
\end{figure*}

\subsection{Analysis}
\textbf{Varying Size of $\overline{C_s}$ and $\overline{C_t}$.} To investigate the performance of our method under different Noisy UniDA settings, we fixed $C_s \cup C_t$ and $C$ in the task A$\rightarrow$D with noise type P20 in the Office dataset and changed the sizes of $\overline{C_t}$ ($\overline{C_s}$ is also changed by $\overline{C_s}=C_s \cup C_t \setminus C \setminus \overline{C_t}$). \Fref{fig:ab_ct} shows the comparison of our method and other methods on different sizes of $\overline{C_t}$. When $|\overline{C_t}|=0$, which is the partial domain adaptation setting with $C_t \subset C_s$, the performance of our method is comparable to that of ETN, which is a partial domain adaptation method. When $|\overline{C_t}|=21$, which is the open-set domain adaptation setting with $C_s \subset C_t$, the performance of our method is comparable to that of DANCE. In the middle of 0 and 21, where $C_s$ and $C_t$ are partially shared, our method outperforms other methods with a large margin. It is also interesting to find that the general UniDA methods, UAN and DANCE, perform poorly when $|\overline{C_t}|=0$ (partial DA) owing to the noisy source samples.

\textbf{Varying Noise Level.} We further explored the effect of label noise on performance. We changed the noise level of the pairflip and symmetric noise from 0 to 0.45 on task A$\rightarrow$D in the Office dataset. \Fref{fig:ab_p} and \Fref{fig:ab_s} show the results and our method achieve high performance in all the settings. It is noticeable that our method is comparable to the state-of-the-art UniDA method (DANCE) when the noise level is 0, which implies that all labels of source samples are clean. Considering the pairflip noise, the damage of the label noise to the performance is large, especially when the noise level is high, but our method still performs better. Considering symmetric noise, although the performance of other methods decreases when the noise level increases, our method is robust to label noise. 

\textbf{Ablation Study.} We further demonstrated the effectiveness of our method by evaluating its variants on the Office dataset with P20 noise. (1) Ours w/o select is the variant without using the small-loss selection by \Eref{eq:select}. (2) Ours w/o div is the variant without using the divergence component in the classification of source samples in \Eref{eq:Ls}. (3) Ours w/o crs is the variant without using the cross-entropy $\mathcal{\tilde{L}}_{crs}$ between the classifiers of target samples in \Eref{eq:Lt}. (4) Ours w/o ent is the variant without using the entropy $\mathcal{\tilde{L}}_{ent}$ of each classifier of target samples in \Eref{eq:Lt}. (5) Ours w/o sep is the variant without separating the divergence between the classifiers as \Eref{eq:Lt}. (6) Ours w/o mini-max is the variant without mini-max training of the generator and classifiers in \Eref{eq:step_b} and \Eref{eq:step_c} to achieve domain alignment. (7) Ours w/ KL is the variant using general symmetric KL-divergence in \Eref{eq:Lt}. As shown in \Tref{tbl:ab}, our method outperforms other variants in all the settings.

\section{Conclusion}
In this paper, we proposed divergence optimization for noisy UniDA. This method uses two classifiers to find clean source samples, reject target private classes, and find important target samples that contribute most to the model's adaptation. We evaluated it on a diverse set of benchmarks, and the proposed method significantly outperformed the current state-of-the-art methods on different setups across various source and target domain pairs.

\subsubsection*{Acknowledgement}
The research was supported by JST ACT-I Grant Number JPMJPR17U5. The research was also partially supported by JSPS KAKENHI Grant Number JP17H06100, Japan.

\begin{table}[t]
\centering
\scalebox{0.9}{
\begin{tabular}{c|ccc|c}
\toprule
Method & \multicolumn{3}{c|}{Office} & 6 Tasks\\
       & D2W & A2D & W2A & Avg \\ \hline \hline
Ours w/o select     &  94.85  &  88.46 & 85.91  & 86.02 \\
Ours w/o div     &  96.27  &  88.18 & 87.08  & 90.14 \\
Ours w/o crs     &  96.06  & 86.88 & 89.63  & 90.30 \\
Ours w/o ent     &  96.30  &  87.60 & 86.30 & 90.45 \\
Ours w/o sep    &  94.78  & 84.89 &  86.71 &  88.53 \\
Ours w/o mini-max   &  96.05  & 88.71 & 81.03  & 87.31 \\
Ours w/ KL    &  96.19  & 86.49 &  87.77 &  89.75 \\
Ours   &  \textbf{96.65}  & \textbf{89.42} & \textbf{90.90} &  \textbf{91.22} \\ \bottomrule
\end{tabular}
}
\vspace{-3pt}
\caption{Ablation study tasks on the Office dataset.}
\vspace{-3pt}

\label{tbl:ab}
\end{table}

\clearpage
{\small
\bibliographystyle{ieee_fullname}
\bibliography{egbib}
}

\end{document}


\title{Supplementary Material: Divergence Optimization \\ for Noisy Universal Domain Adaptation}

\author{Qing Yu\affmark[1,2]\ \ \ \ Atsushi Hashimoto\affmark[2]\ \ \ \ Yoshitaka Ushiku\affmark[2]\\
 \affmark[1]The University of Tokyo\ \ \ \ \affmark[2]OMRON SINIC X Corporation\\ 
{\tt\small yu@hal.t.u-tokyo.ac.jp}\ \ \ \ {\tt\small \{atsushi.hashimoto,yoshitaka.ushiku\}@sinicx.com}
}

\maketitle

\section{Experiments on Toy Datasets.}
\begin{figure*}[t]
     \centering
     \begin{subfigure}[b]{0.33\textwidth}
         \centering
         \includegraphics[width=1\textwidth]{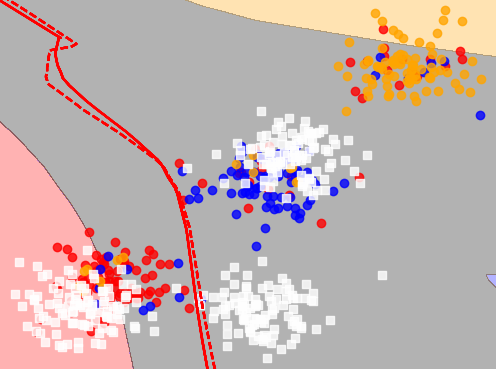}
         \caption{Source Only}
         \label{fig:toy_so}
     \end{subfigure}
     \hfill
     \begin{subfigure}[b]{0.33\textwidth}
         \centering
         \includegraphics[width=1\textwidth]{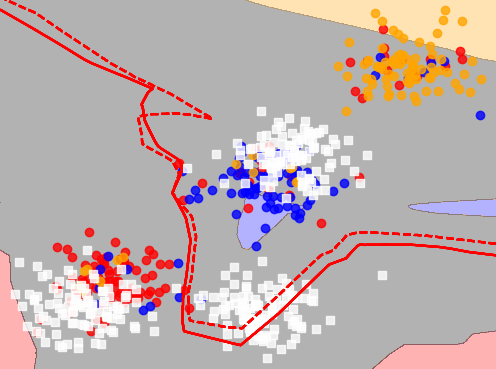}
         \caption{Ours w/o select}
        \label{fig:toy_sel}
     \end{subfigure}
     \hfill
     \begin{subfigure}[b]{0.33\textwidth}
         \centering
         \includegraphics[width=1\textwidth]{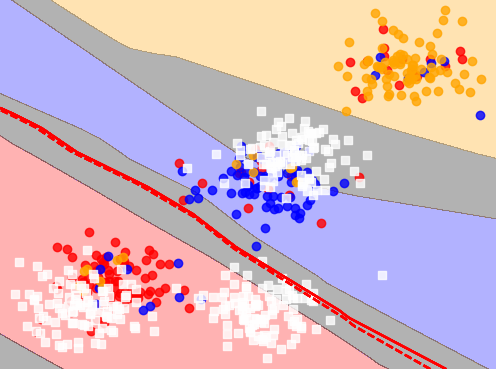}
         \caption{Ours w/o sep.}
         \label{fig:toy_sep}
     \end{subfigure}
     \centering
     \begin{subfigure}[b]{0.33\textwidth}
         \centering
         \includegraphics[width=1\textwidth]{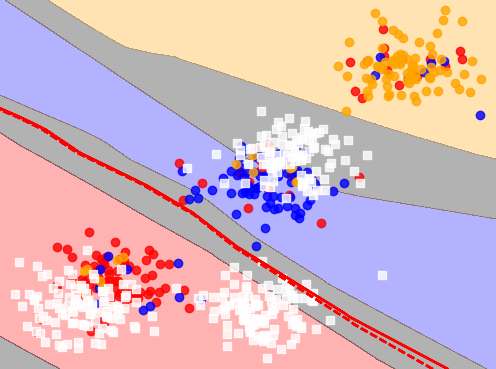}
         \caption{Ours w/ KL.}
         \label{fig:toy_kl}
     \end{subfigure}
     \begin{subfigure}[b]{0.33\textwidth}
         \centering
         \includegraphics[width=1\textwidth]{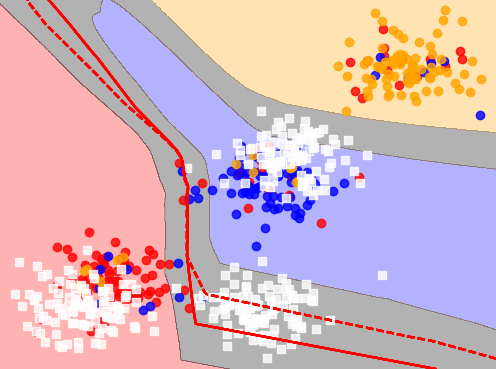}
         \caption{Ours.}
         \label{fig:toy_ours}
     \end{subfigure}
    \caption{(Best viewed in color.) Visualization of the toy problem. Red, blue and orange points indicate the three classes of the source samples, where the orange class is the source private class. The labels of source samples are corrupted with 20\% symmetric noise. White points represent target samples, and the target samples at the right bottom of each figure are target private samples. The dashed and normal lines are two decision boundaries in our method for identifying the red class. The pink, light blue and light yellow regions are where the results of both classifiers are class red, blue and orange, respectively. The gray regions are detected as the target private class having large divergence by Eq. (15).}
    \label{fig:toy}
\end{figure*}

We observed the behavior of the proposed method on a toy dataset, where we used scikit-learn to generate isotropic Gaussian blobs as the source samples and the target samples. The goal of the experiment was to observe the learned classifiers’ boundaries. For the source samples, we generated three clusters of blobs and labeled them as red, blue and orange, respectively. Some label noise is also introduced in source samples. Target common samples were generated around the distribution of the red and blue source samples, where target private samples were generated far from the source samples. We generated 300 source and target samples per class as the training samples. The model consists of a 3-layered fully-connected network for a feature generator and 3-layered fully-connected networks for classifiers. In \Fref{fig:toy}, we visualized the learned decision boundary of the proposed method and its variants as follows:

\textbf{Source Only} 
The method that training with source samples only with Eq. (7), which means the final objectives are as follows:
\setcounter{equation}{15}
\begin{equation}
  \mymin_{G,F_1,F_2} \mathcal{L}_{sup}(D_s).
\end{equation}

\textbf{Ours w/o select} 
The method that training without using the small-loss selection by Eq. (9), which means the objectives of Step A-1 are as follows:
\begin{equation}
  \mymin_{G,F_1,F_2} \mathcal{L}_s(D_s).
\end{equation}

\textbf{Ours w/o sep} 
The method that training without separating the divergence between the classifiers as Eq. (11), which means the following equation is used instead of Eq. (11):
\begin{equation}
  \mathcal{L}_{t}(D_t) = 0.
\end{equation}

\textbf{Ours w/ KL} 
The method that training with using general symmetric KL-divergence in Eq. (11), which means the following equation is used instead of Eq. (11):
\begin{equation}
\label{eq:Lt}
   \mathcal{L}_{t}(D_t) = \mathcal{\tilde{L}}_{SKLD}(D_t)={\frac{1}{N}}\sum_{i=1}^{N}\mathcal{\tilde{L}}_{crs}(D_t)-{\frac{1}{N}}\sum_{i=1}^{N}\mathcal{\tilde{L}}_{ent}(D_t)
\end{equation}

Regarding Source Only and Ours w/o select shown in \Fref{fig:toy_so} and \Fref{fig:toy_sel}, a large region is detected as the target private class having large $\mathcal{L}_{crs}$ due to the label noise of the source samples. As for Ours w/o sep and Ours w/ KL shown in \Fref{fig:toy_sep} and \Fref{fig:toy_kl}, though they are not effected by the label noise, they cannot detect target private samples well because their loss function is not suitable for separate the target common and target private classes. Compared to other variants, the proposed method shown in \Fref{fig:toy_ours} achieves the best performance. In our proposed method, the classifiers are not effected by the noisy source samples, and target private samples exist in the area having large divergence, where our proposed method attempted to increase the divergence on these target private samples and decrease that on target common samples. The code of the toy problem is provided in the supplementary.

\section{Sensitivity to hyper-parameters.}
\begin{figure*}[t]
     \centering
     \begin{subfigure}[b]{0.33\textwidth}
         \centering
         \includegraphics[width=1.1\textwidth]{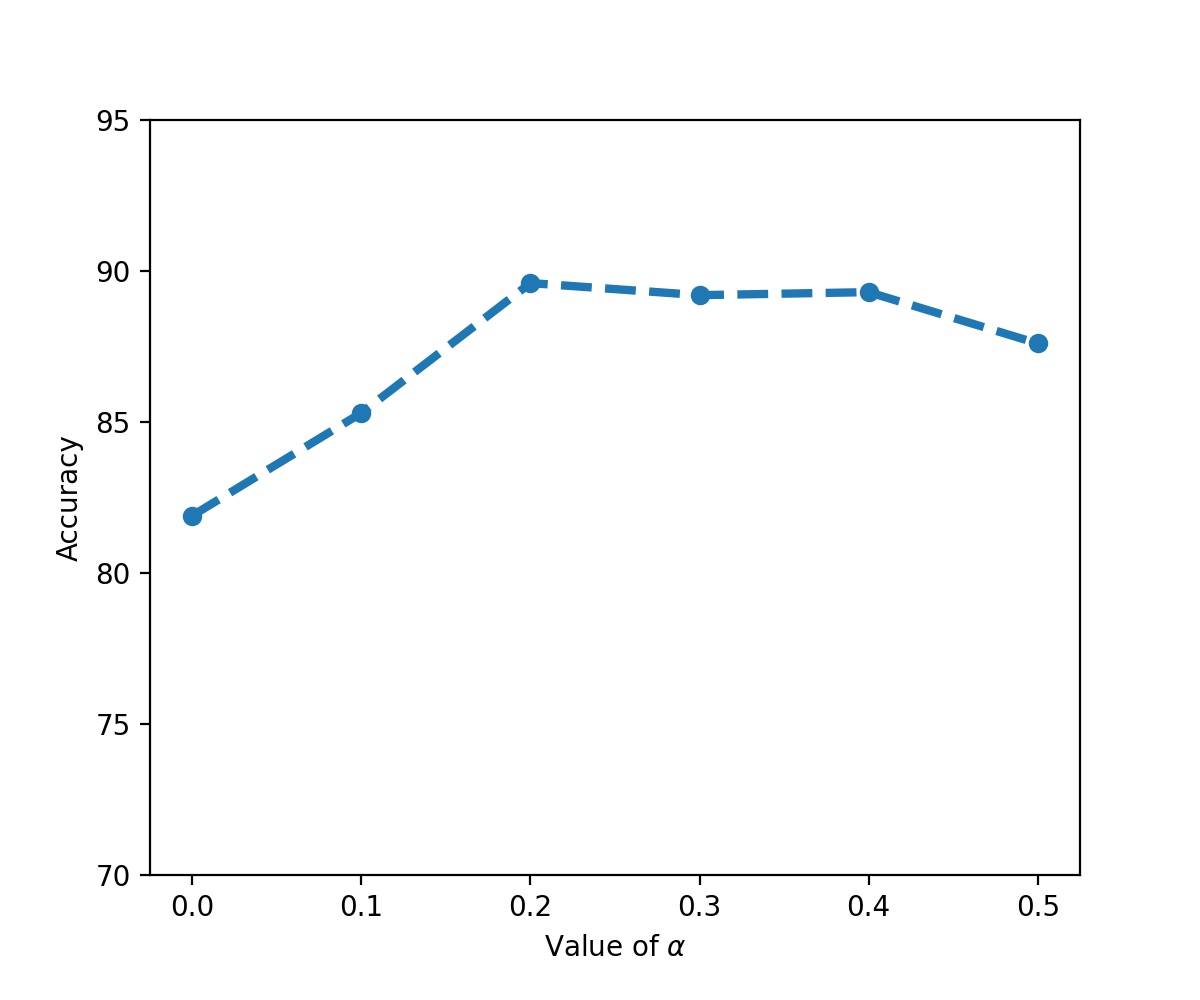}
         \caption{Accuracy w.r.t. value of $\alpha$ in Eq. (9).}
         \label{fig:ab_alpha}
     \end{subfigure}
     \hfill
     \begin{subfigure}[b]{0.33\textwidth}
         \centering
         \includegraphics[width=1.1\textwidth]{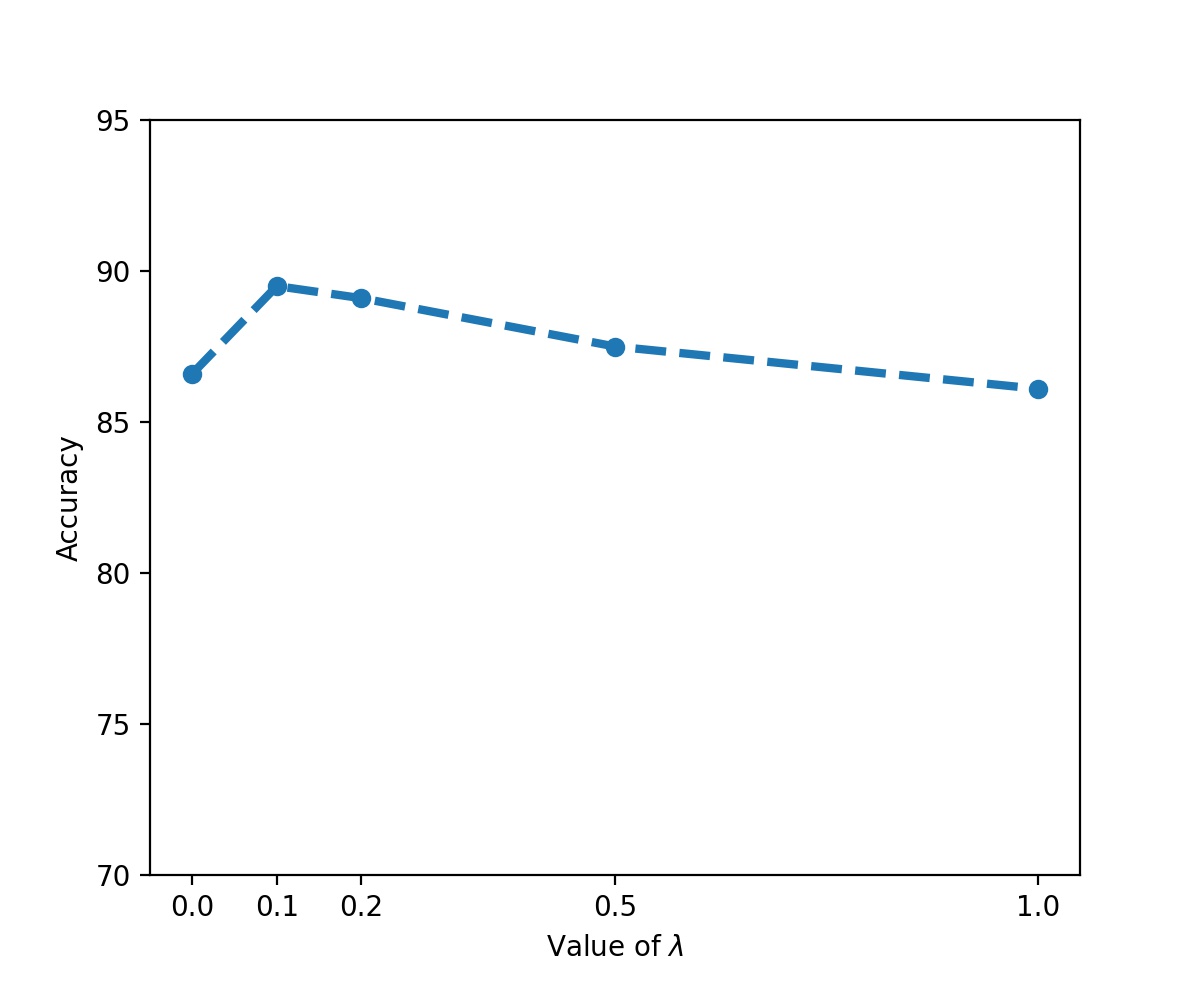}
         \caption{Accuracy w.r.t. value of $\lambda$ in Eq. (8).}
        \label{fig:ab_lambda}
     \end{subfigure}
     \hfill
     \begin{subfigure}[b]{0.33\textwidth}
         \centering
         \includegraphics[width=1.1\textwidth]{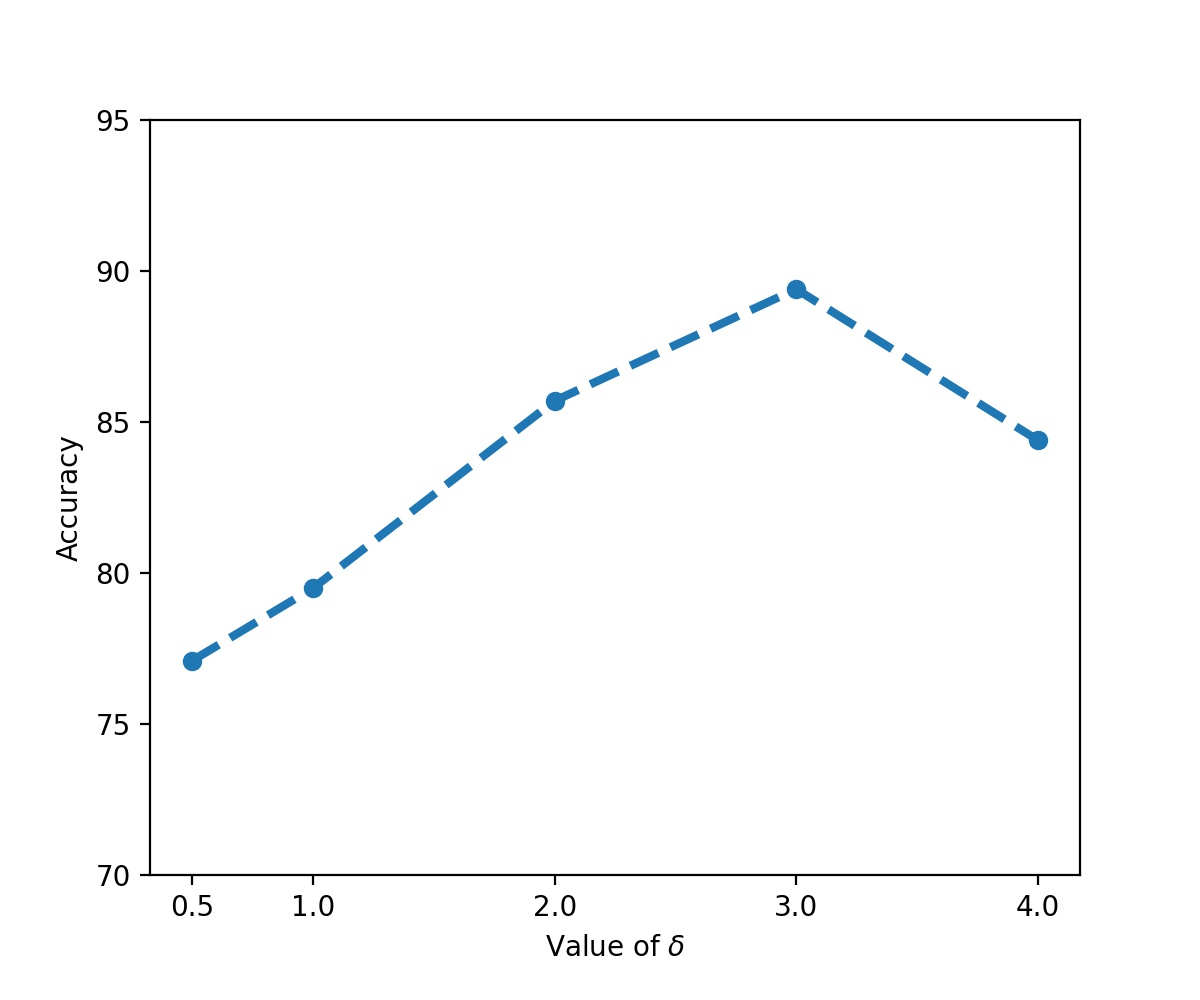}
         \caption{Accuracy w.r.t. value of $\delta$ in Eq. (11).}
         \label{fig:ab_delta}
     \end{subfigure}
     \centering
     \begin{subfigure}[b]{0.33\textwidth}
         \centering
         \includegraphics[width=1.1\textwidth]{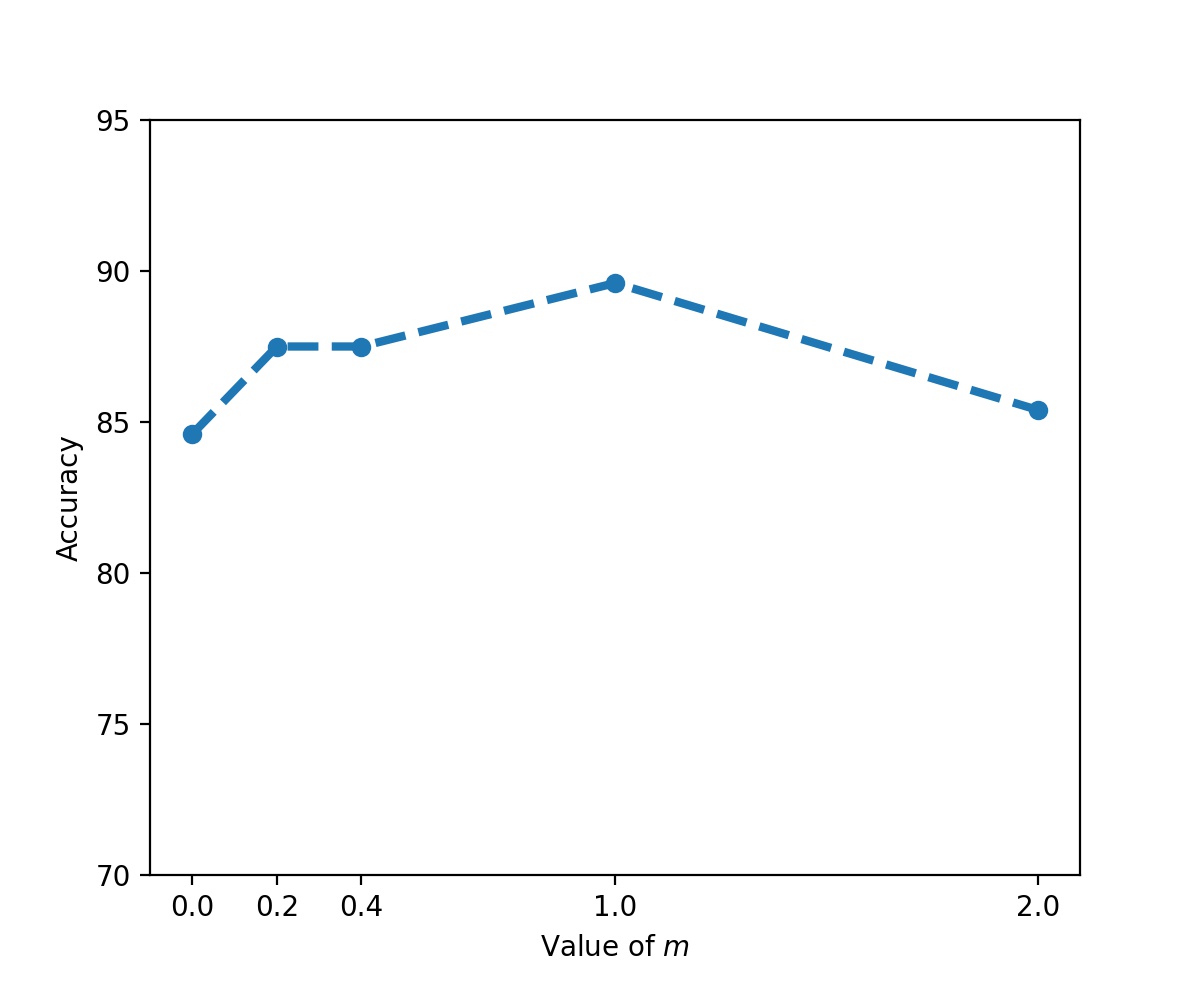}
         \caption{Accuracy w.r.t. value of $m$ in Eq. (11).}
         \label{fig:ab_m}
     \end{subfigure}
     \begin{subfigure}[b]{0.33\textwidth}
         \centering
         \includegraphics[width=1.1\textwidth]{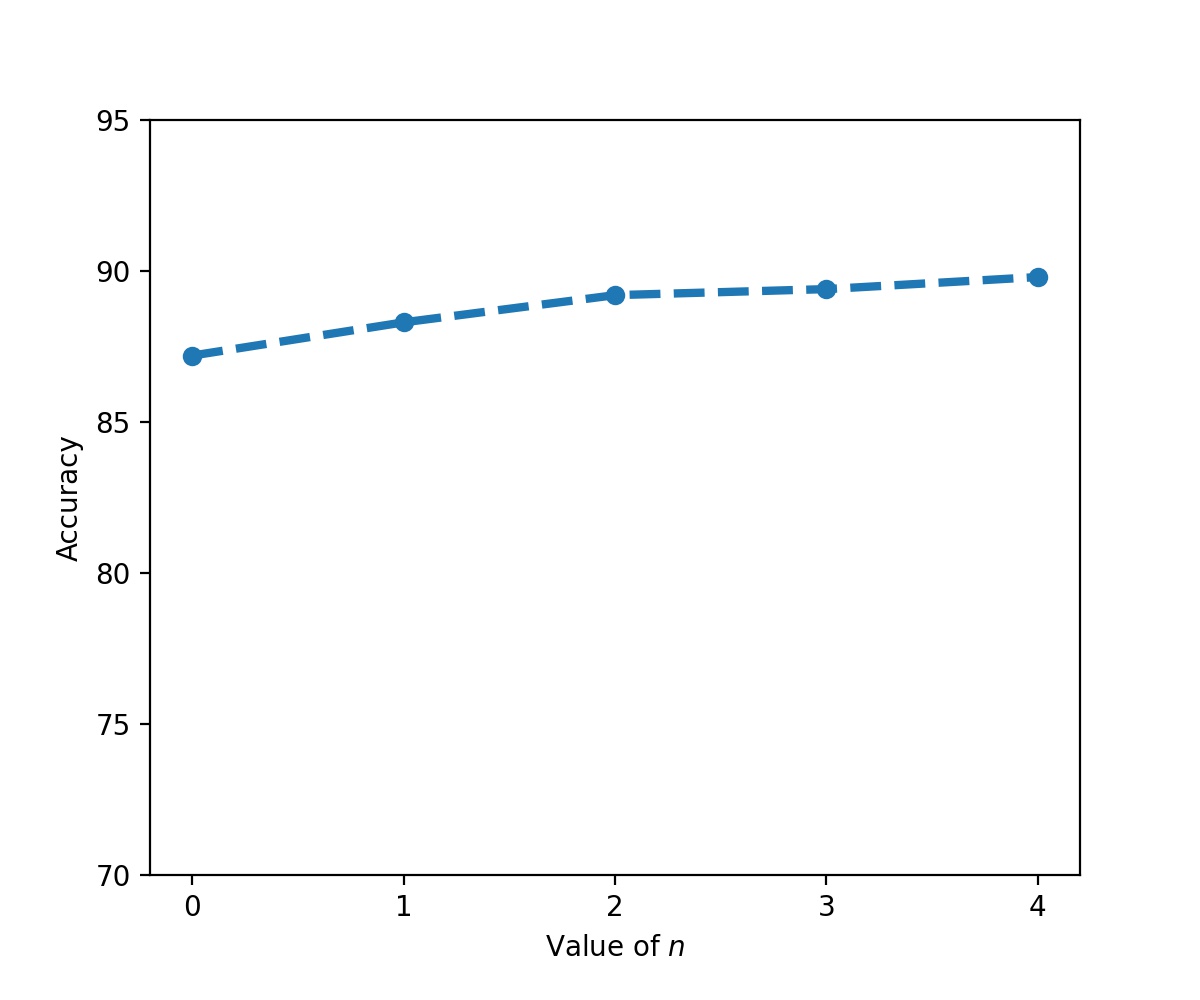}
         \caption{Accuracy w.r.t. value of $n$ in Step C.}
         \label{fig:ab_n}
     \end{subfigure}
    \caption{Analysis of the sensitivity to hyper-parameters in task A$\rightarrow$D with P20 noise.}
    \label{fig:ab_hp}
\end{figure*}

\begin{figure*}[t]
     \centering
     \begin{subfigure}[b]{0.33\textwidth}
         \centering
         \includegraphics[width=1.1\textwidth]{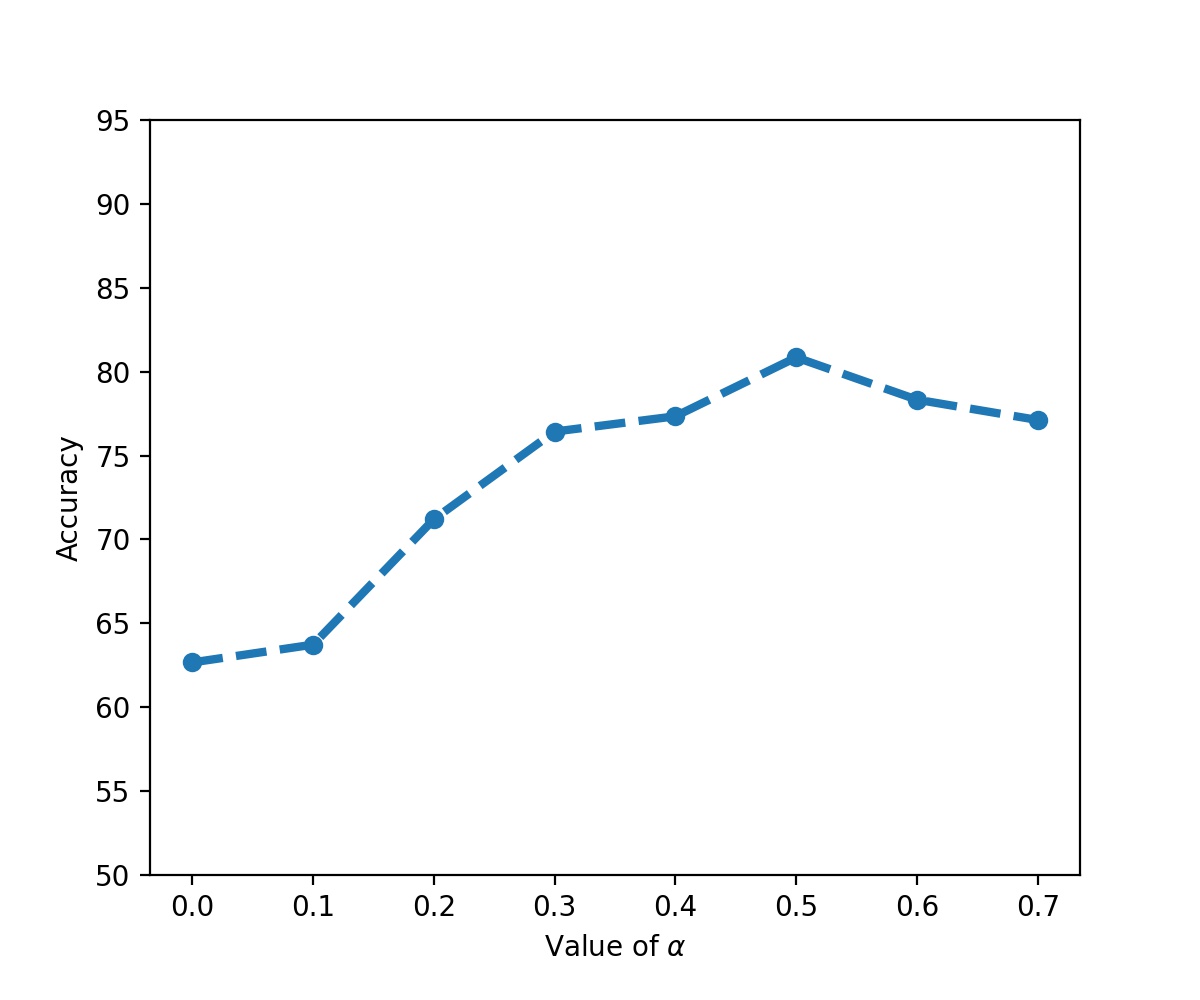}
         \caption{Accuracy w.r.t. value of $\alpha$ in Eq. (9).}
         \label{fig:ab_alpha_2}
     \end{subfigure}
     \hfill
     \begin{subfigure}[b]{0.33\textwidth}
         \centering
         \includegraphics[width=1.1\textwidth]{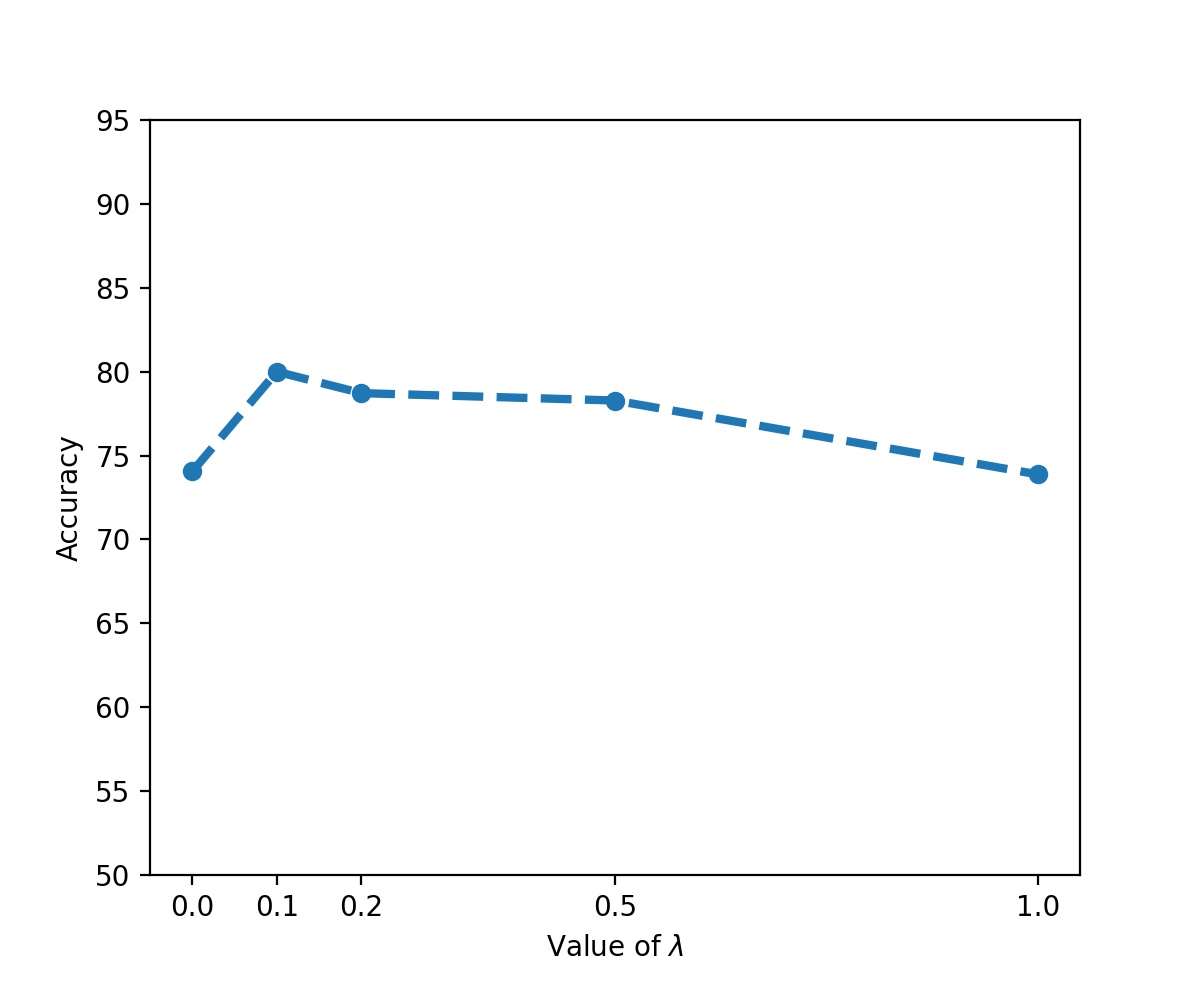}
         \caption{Accuracy w.r.t. value of $\lambda$ in Eq. (8).}
        \label{fig:ab_lambda_2}
     \end{subfigure}
     \hfill
     \begin{subfigure}[b]{0.33\textwidth}
         \centering
         \includegraphics[width=1.1\textwidth]{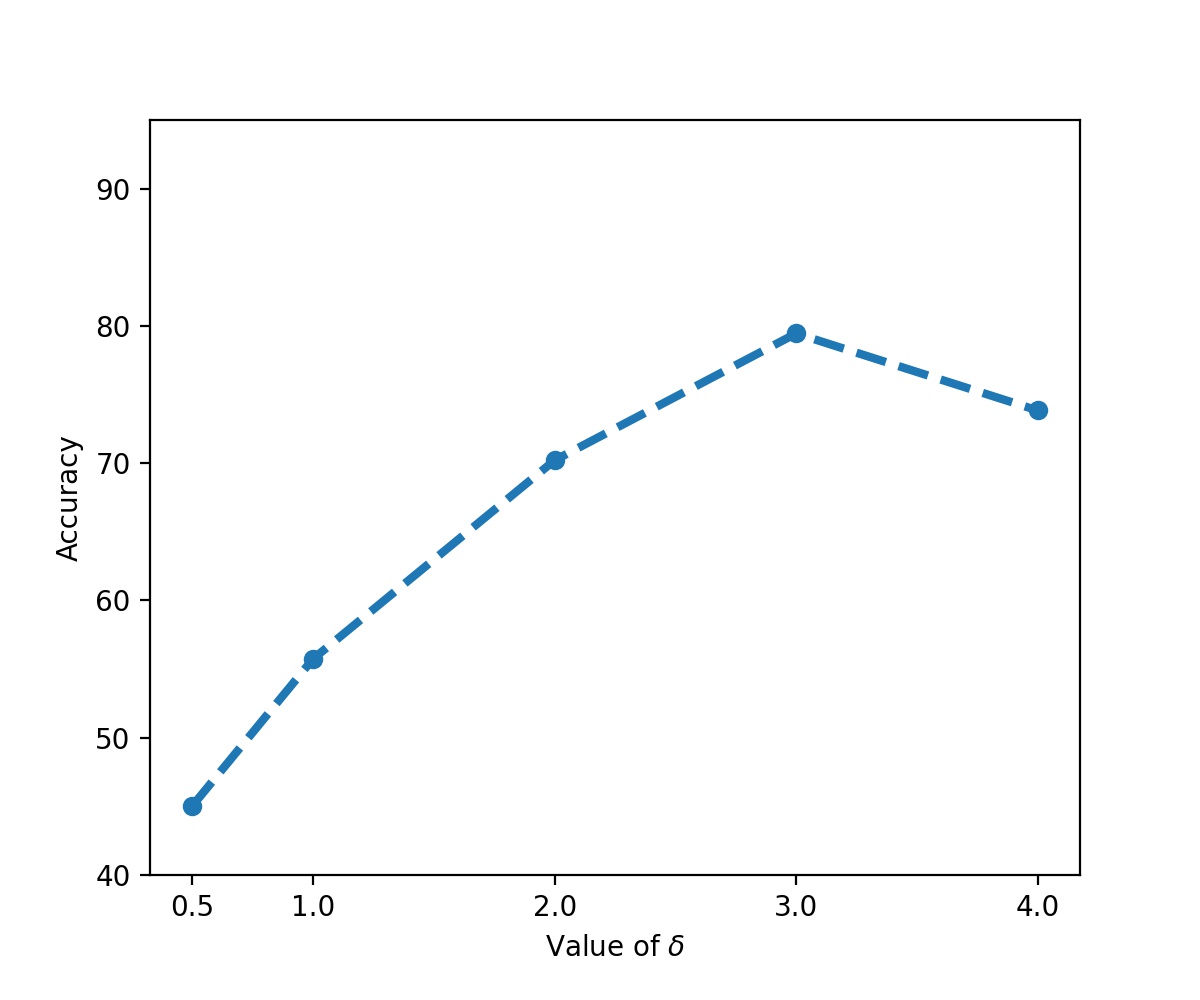}
         \caption{Accuracy w.r.t. value of $\delta$ in Eq. (11).}
         \label{fig:ab_delta_2}
     \end{subfigure}
     \centering
     \begin{subfigure}[b]{0.33\textwidth}
         \centering
         \includegraphics[width=1.1\textwidth]{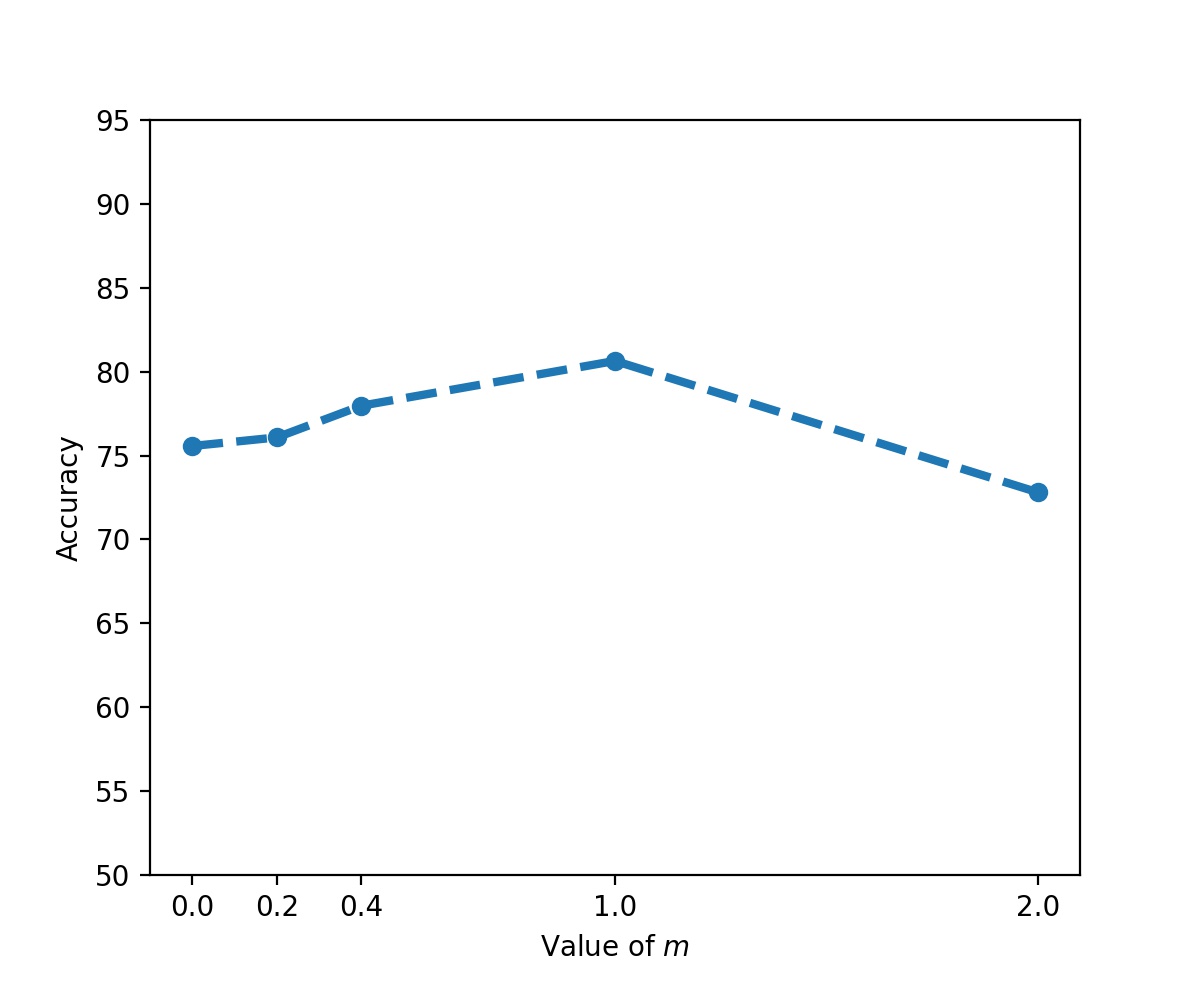}
         \caption{Accuracy w.r.t. value of $m$ in Eq. (11).}
         \label{fig:ab_m_2}
     \end{subfigure}
     \begin{subfigure}[b]{0.33\textwidth}
         \centering
         \includegraphics[width=1.1\textwidth]{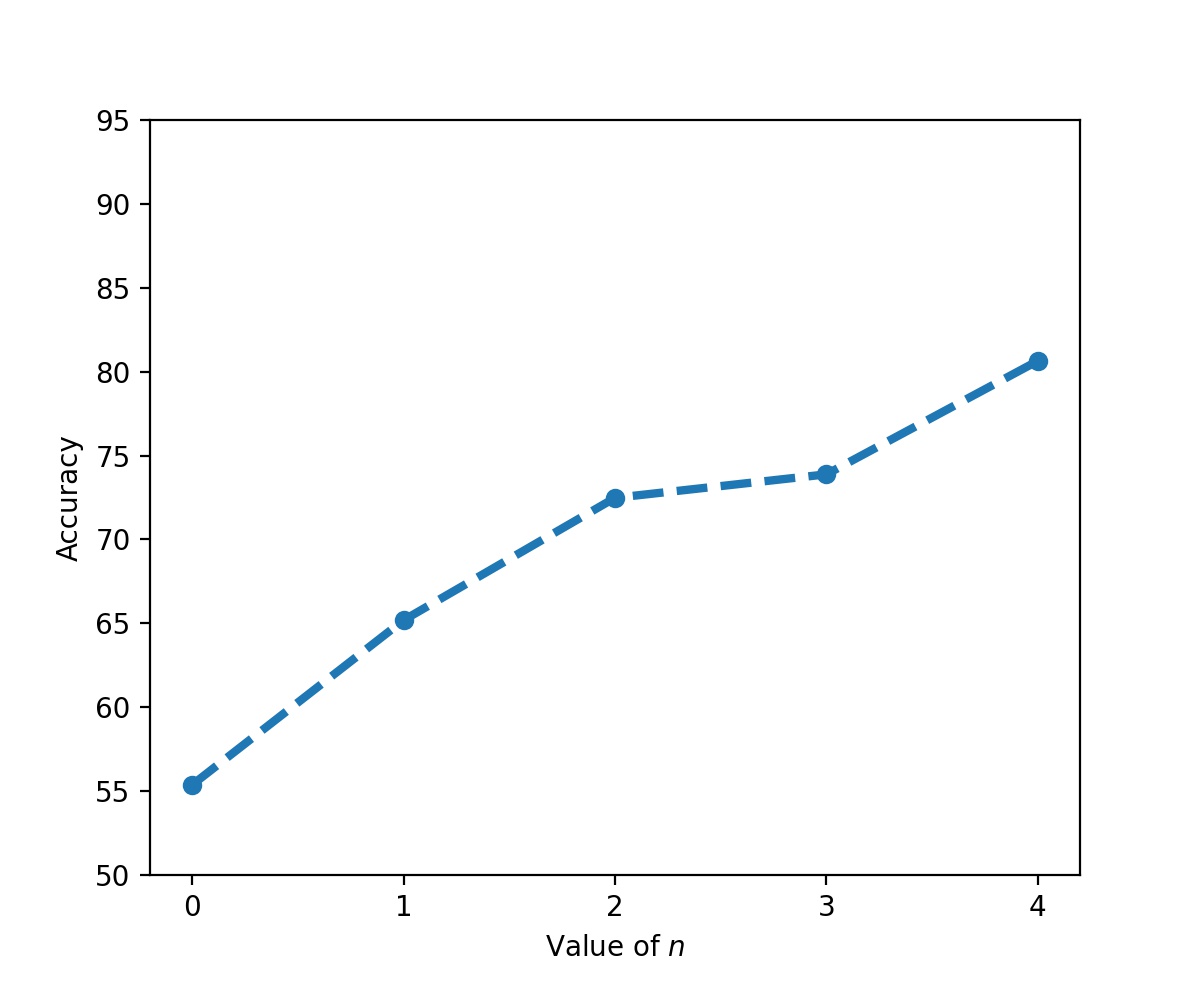}
         \caption{Accuracy w.r.t. value of $n$ in Step C.}
         \label{fig:ab_n_2}
     \end{subfigure}
    \caption{Analysis of the sensitivity to hyper-parameters in task D$\rightarrow$W with S45 noise.}
    \label{fig:ab_hp_2}
\end{figure*}

In \Fref{fig:ab_hp} and \Fref{fig:ab_hp_2}, we show the sensitivity to hyper-parameters on task A$\rightarrow$D with P20 noise and D$\rightarrow$W with S45 noise in the Office dataset, respectively. Regarding $\alpha$, which controls the number of source samples ignored in each mini-batch, we can see that when no samples are dropped ($\alpha=0$), the performance could deteriorate due to the label noise. Though the true noise rate is 0.2 and 0.45, dropping more samples with larger $\alpha$ still could achieve good performance. As for $\lambda$, which controls the weight of divergence loss on source samples, $\lambda=0.1$ shows the best results in both settings. Regarding $\delta$ and $m$, which control divergence separation on target samples, deciding $\delta$ as $\log |C_s| \approx 3$ works well, and setting $m$ around $1$ achieves best performances. As for $n$, which is the time of repeating Step C during the training process, $n>1$ shows similar results in the setting of P20 A$\rightarrow$D, and $n=4$ shows the best results in the setting of S45 D$\rightarrow$W.

